\documentclass[10pt,twocolumn,letterpaper]{article}

\usepackage{cvpr}              %

\usepackage[dvipsnames]{xcolor}
\usepackage{graphicx}
\usepackage{balance}

\newcommand{\CSWildPlaces}{CS-Wild-Places}

\newcommand{\HOTFormerLoc}{HOTFormerLoc}
\newcommand{\carriertoken}{relay token}

\definecolor{cvprblue}{rgb}{0.21,0.49,0.74}
\usepackage[pagebackref,breaklinks,colorlinks,allcolors=cvprblue]{hyperref}

\def\paperID{6081} %
\def\confName{CVPR}
\def\confYear{2025}

\title{\HOTFormerLoc: Hierarchical Octree Transformer \\for Versatile Lidar Place Recognition Across Ground and Aerial Views}

\author{
    Ethan Griffiths$^{1,2}$
    \quad{}Maryam Haghighat$^{1}$\quad{}Simon Denman$^{1}$\quad{}Clinton Fookes$^{1}$\quad{}Milad Ramezani$^{2}$\\
    $^1$Queensland University of Technology (QUT)\quad{}$^2$CSIRO Robotics, Data61, CSIRO\\
    $^1${\tt\small \{maryam.haghighat, s.denman, c.fookes\}@qut.edu.au
    }\\
    $^2${\tt\small \{ethan.griffiths, milad.ramezani\}@data61.csiro.au\
    }
}
\begin{document}
\maketitle
\begin{abstract}

We present \HOTFormerLoc, a novel and versatile \textbf{H}ierarchical \textbf{O}ctree-based \textbf{T}rans\textbf{F}ormer, for large-scale 3D place recognition in both ground-to-ground and ground-to-aerial scenarios across urban and forest environments. 
We propose an octree-based multi-scale attention mechanism that captures spatial and semantic features across granularities. To address the variable density of point distributions from spinning lidar, we present cylindrical octree attention windows to reflect the underlying distribution during attention.
We introduce relay tokens to enable efficient global-local interactions and multi-scale representation learning at reduced computational cost.
Our pyramid attentional pooling then synthesises a robust global descriptor for end-to-end place recognition in challenging environments. 
In addition, we introduce \CSWildPlaces
, a novel 3D cross-source dataset featuring point cloud data from aerial and ground lidar scans captured in dense forests. Point clouds in~\CSWildPlaces~contain representational gaps and distinctive attributes such as varying point densities and noise patterns, making it a challenging benchmark for cross-view localisation in the wild.
HOTFormerLoc achieves a top-1 average recall improvement of 5.5\% -- 11.5\% on the \CSWildPlaces~benchmark. Furthermore, it consistently outperforms SOTA 3D place recognition methods, with an average performance gain of 4.9\% on well-established urban and forest datasets. The code and \CSWildPlaces~benchmark is available at \hyperlink{https://csiro-robotics.github.io/HOTFormerLoc}{https://csiro-robotics.github.io/HOTFormerLoc}.
\vspace{-7mm}
\end{abstract}

\section{Introduction}
\label{sec:intro}

\begin{figure}[t]
  \centering
   \includegraphics[width=0.99\linewidth]{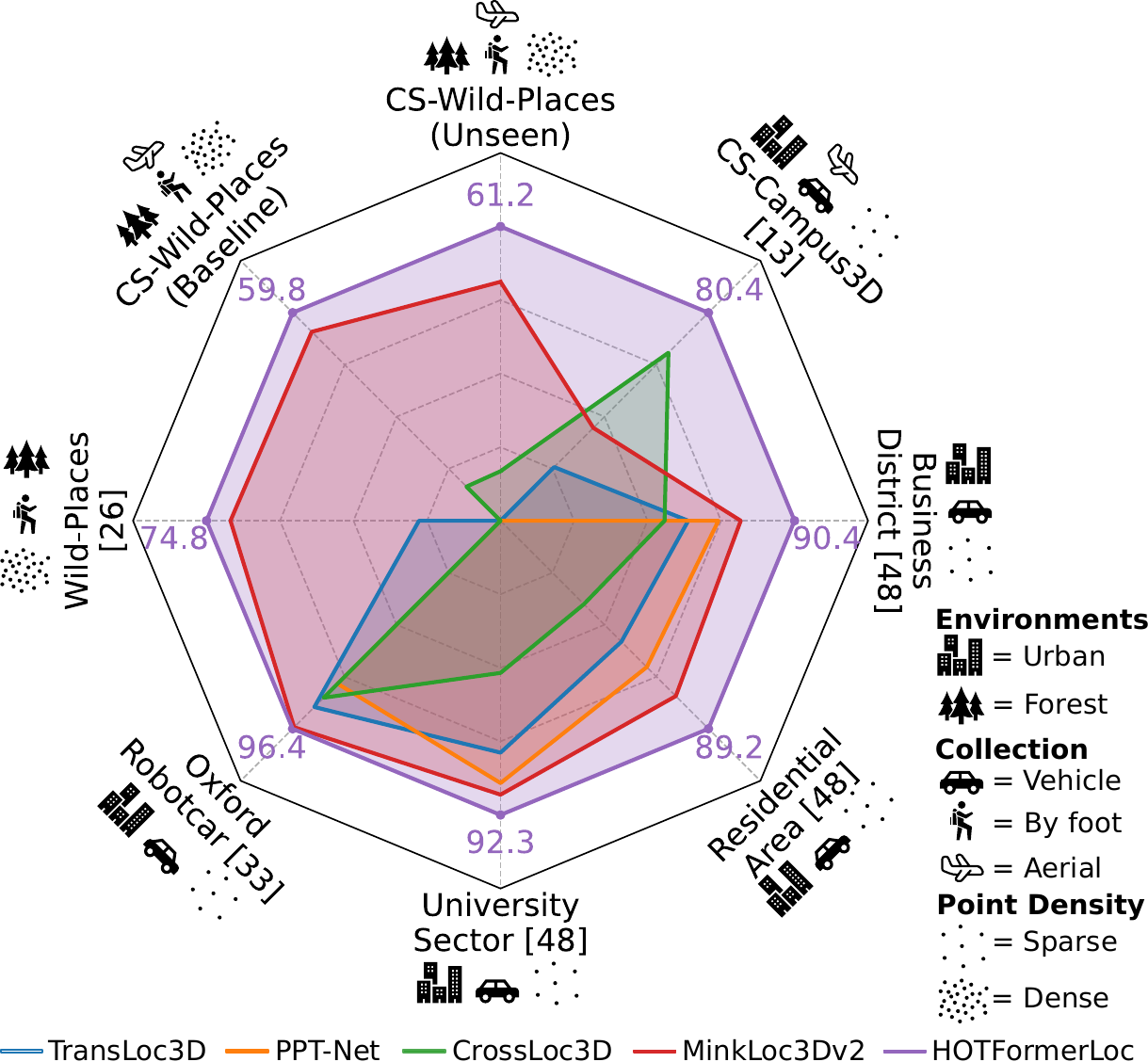}
   \vspace{-2mm}
   \caption{\small{\HOTFormerLoc~achieves SOTA performance across a suite of LPR benchmarks with diverse environments, varying viewpoints, and different point cloud densities. }}
   \label{fig:hero_fig}
   \vspace{-5mm}
\end{figure}

End-to-end Place Recognition (PR) has gained significant attention in computer vision and robotics for its capacity to convert sensory data—such as lidar scans or images—into compact embeddings that effectively capture distinctive scene features. During inference, these embeddings enable the system to match a query frame to stored data, allowing it to recognise previously visited locations. This capability is particularly crucial in autonomous navigation, where reliable PR helps locate the vehicle within a prior map for global localisation~\cite{yinSurveyGlobalLiDAR2024}. When integrated with metric localisation in Simultaneous Localisation and Mapping (SLAM), it helps to reduce drift, enhance map consistency and improve long-term navigation~\cite{yinGeneralPlaceRecognition2022}.

Despite considerable advancements,~Visual Place Recognition (VPR)~\cite{zhangVisualPlaceRecognition2021} struggles with robustness against variations in appearance, season, lighting, and viewpoint, especially within large-scale urban and forested environments, where orthogonal-view scenarios can exacerbate the challenge.~This work specifically addresses the problem of Lidar Place Recognition (LPR) in diverse conditions. While recent LPR methods have achieved notable progress, their primary focus has been on urban settings using ground-to-ground view descriptors, leaving cross-view lidar recognition, particularly in natural environments like forests, largely under-explored.~Natural environments present complexities such as dense occlusions, a scarcity of distinctive features, long-term structural changes, and perceptual aliasing, complicating reliable PR. Further, when relying on sparse point clouds—such as 4096 points, a standard for many LPR baselines—performance declines sharply in forest areas as the sparse points lack the necessary spatial resolution and semantic content.

To address these challenges, we propose a versatile hierarchical octree-based transformer for large-scale LPR in ground-to-ground and ground-to-aerial scenarios, capable of handling diverse point densities in urban and forest environments.~Our approach enables multi-scale feature interaction and selection via compact proxies in a hierarchical attention mechanism, circumventing the prohibitive cost of full attention within large point clouds. This design captures global context across multiple granularities—especially valuable in areas with occlusions and limited distinctive features.
We introduce point serialisation with cylindrical octree attention windows to align attention with point distributions from common spinning lidar. This approach is critical for managing point clouds in cluttered environments and mitigates the risk of over-confident predictions arising from varying point densities.
We demonstrate the versatility of our method with SOTA performance on a diverse suite of LPR benchmarks (see \cref{fig:hero_fig}).
The contributions of this paper are as follows: 
 \begin{itemize}
     \item \textit{\HOTFormerLoc}, a novel octree-based transformer with \textbf{hierarchical attention} that efficiently relays long-range contextual information across multiple scales.
     \item \textbf{Cylindrical Octree Attention} to better represent the variable density of point clouds captured by spinning lidar— denser near the sensor and sparser at a distance.
     \item \textbf{Pyramid Attentional Pooling} to adaptively select and aggregate local features from multiple scales into global descriptors for end-to-end PR.
     \item \textbf{\CSWildPlaces}, the first large-scale ground-aerial LPR dataset in unstructured, forest environments. 
 \end{itemize}

\section{Related Works}
\label{sec:relatedworks}

\noindent
\textbf{Lidar Place Recognition:}
LPR methods have evolved from handcrafted approaches,~\eg~\cite{kimScanContextEgocentric2018,he2016m2dp}, toward deep learning methods trained in a metric-learning framework.
Previous approaches~\cite{uyPointNetVLADDeepPoint2018,komorowskiMinkLoc3DPointCloud2021,komorowskiImprovingPointCloud2022,vidanapathiranaLoGG3DNetLocallyGuided2022,huiPyramidPointCloud2021,xuTransLoc3DPointCloud2022} encode point clouds into local descriptors using backbones built on PointNet~\cite{qiPointNetDeepLearning2017}, sparse 3D CNNs~\cite{choy4DSpatioTemporalConvNets2019,tangSearchingEfficient3D2020}, or transformers~\cite{vaswaniAttentionAllYou2017b}.
These descriptors are then aggregated into global embeddings through pooling methods like NetVLAD~\cite{arandjelovicNetVLADCNNArchitecture2016}, GeM~\cite{radenovicFineTuningCNNImage2019}, or second-order pooling~\cite{vidanapathiranaLoGG3DNetLocallyGuided2022}, facilitating end-to-end PR.

\noindent
\textbf{Cross-source Lidar Place Recognition:}
To address LPR for cross-source/cross-view matching,~\cite{xumingSemanticMapsCrossview2022} creates 2.5D semantic maps from both viewpoints to co-register point clouds. CrossLoc3D~\cite{guanCrossLoc3DAerialGroundCrossSource2023} learns multi-scale features with sparse 3D CNNs and refines them into a canonical feature space inspired by diffusion~\cite{hoDenoisingDiffusionProbabilistic2020}. While effective on ground-aerial LPR scenarios, it slightly underperforms on single-source benchmarks. GAPR~\cite{jieHeterogeneousDeepMetric2023} uses sparse 3D CNN with PointSoftTriplet, a modified soft margin triplet loss from~\cite{hu2018cvm}, and an attention-based overlap loss to enhance consistency by focusing on high-overlap regions.

\noindent
\textbf{Point Cloud Transformers:} 
Following the success of transformers in NLP~\cite{vaswaniAttentionAllYou2017b} and computer vision~\cite{dosovitskiyImageWorth16x162021a, liuSwinTransformerHierarchical2021a}, point cloud transformers have gained traction for 3D representation learning. %
However, initial architectures~\cite{guoPCTPointCloud2021,zhaoPointTransformer2021} are hindered by quadratic memory costs, restricting application. Efforts to manage this use vector attention~\cite{zhaoPointTransformer2021, wuPointTransformerV22022} but involve computationally heavy sampling and pooling steps.
Newer architectures have taken cues from Swin Transformer~\cite{liuSwinTransformerHierarchical2021a}, restricting attention to non-overlapping local windows. However, the sparse nature of point clouds creates parallelisation challenges due to varied window sizes. Various solutions have been proposed to alleviate this issue~\cite{yangSwin3DPretrainedTransformer2023a,fanEmbracingSingleStride2022,sunSWFormerSparseWindow2022,laiStratifiedTransformer3D2022}, at the cost of bulky implementations.

Serialisation-based transformers overcome these inefficiencies by converting point clouds into ordered sequences, enabling structured attention over equally sized windows. FlatFormer~\cite{liuFlatFormerFlattenedWindow2023} employs window-based sorting for speed-ups, while OctFormer~\cite{wangOctFormerOctreebasedTransformers2023} leverages octrees with Z-order curves~\cite{mortonComputerOrientedGeodetic1966} for efficient dilated attention and increased receptive field. PointTransformerV3~\cite{wuPointTransformerV32024} introduces randomised space-filling curves~\cite{peanoCourbeQuiRemplit1890} to improve scalability across benchmarks. However, even with such advancements, these efficient transformers are limited by reduced receptive field, which restricts global context learning.

\begin{figure*}[t]
  \centering
  \begin{subfigure}{0.69\linewidth}
    \includegraphics[width=\textwidth]{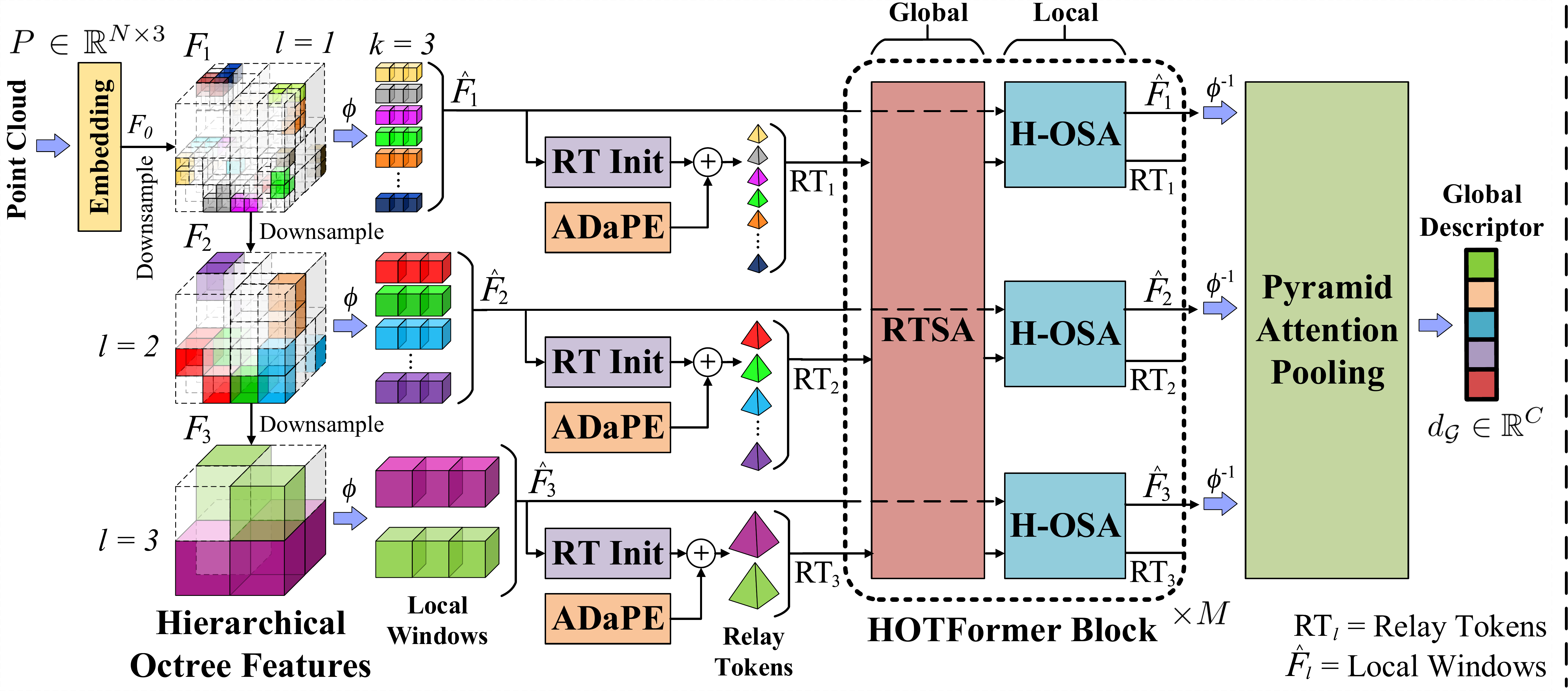}
    \caption{HOTFormerLoc Architecture}
    \label{fig:hotformer_diagram}
  \end{subfigure}
  \begin{subfigure}{0.30\linewidth}
    \includegraphics[width=\textwidth]{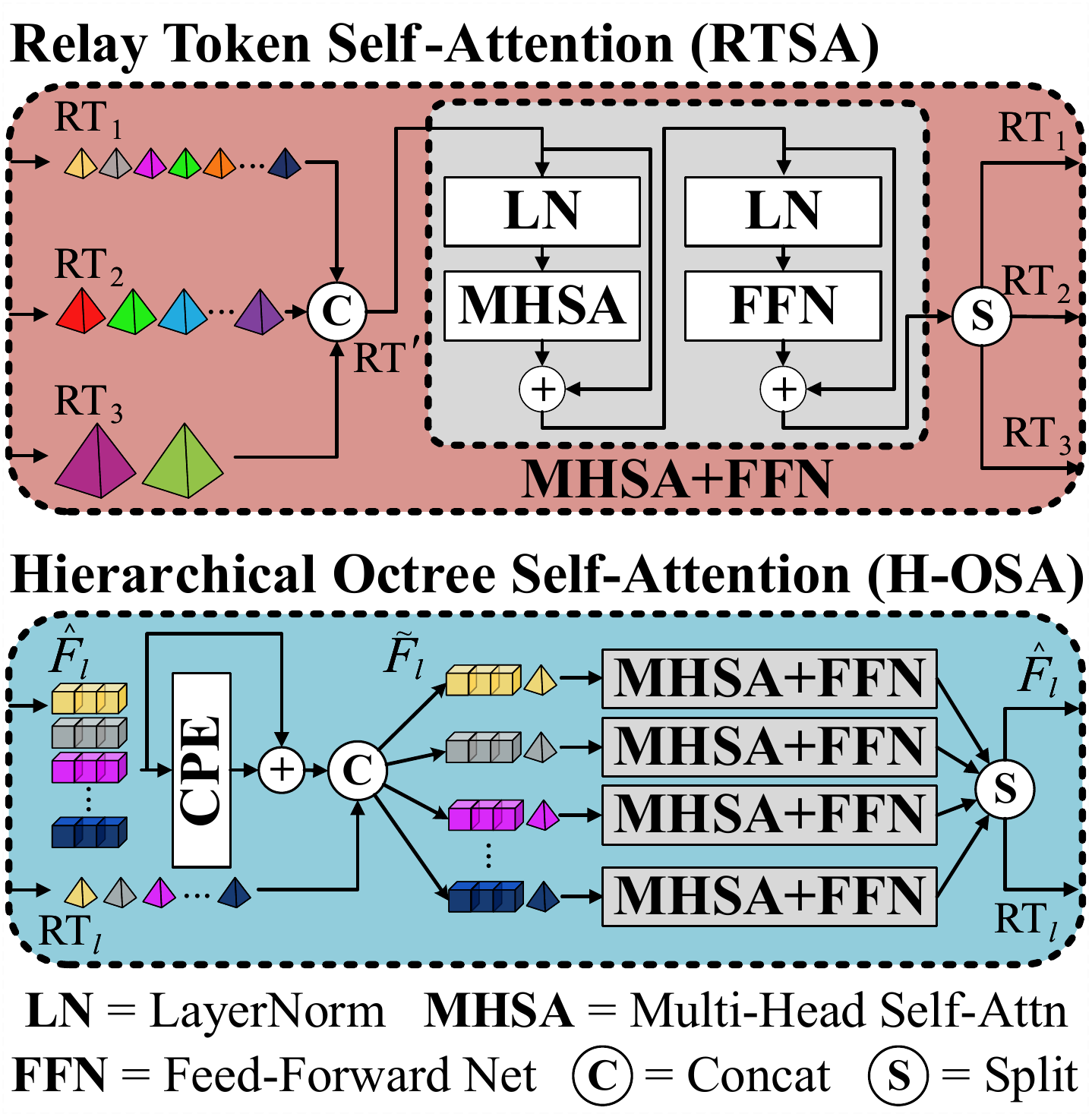}
    \caption{RTSA and H-OSA Blocks}
    \label{fig:hierarchical_attention}
  \end{subfigure}
  \vspace{-2mm}
  \caption{\small{(a) We use an octree to guide a hierarchical feature pyramid $F$, which is tokenised and partitioned into local attention windows $\hat{F}_l$ of size $k$ ($k=3$ in this example). We introduce a set of relay tokens $RT_l$ to represent local regions at each level, and process both local and \carriertoken{}s in a series of HOTFormer blocks. Pyramid attention pooling then aggregates multi-scale features into a single global descriptor. (b) HOTFormer blocks consist of relay token self-attention (RTSA) to induce long distance multi-scale interactions, and hierarchical octree self-attention (H-OSA) to refine local features and propagate global contextual cues learned by the relay tokens.}}
  \label{fig:hotformer_overview}
  \vspace{-5mm}
\end{figure*}
\noindent
\textbf{Hierarchical Attention:}
Driven by limitations in ViT approaches with single-scale tokens, interest in hierarchical attention mechanisms has grown.
Recent image transformers~\cite{wangPyramidVisionTransformer2021a,liuSwinTransformerHierarchical2021a}, generate multi-scale feature maps suited for tasks like segmentation and object detection, but lack global interactions due to partitioning.

To expand the receptive field, CrossViT~\cite{chenCrossViTCrossAttentionMultiScale2021a} introduces cross-attention to fuse tokens from multiple patch sizes, while Focal Transformer~\cite{yangFocalAttentionLongRange2021a} adapts token granularity based on pixel distance. Twins~\cite{chuTwinsRevisitingDesign2021a} and EdgeViT~\cite{panEdgeViTsCompetingLightweight2022} combine window attention with global attention to model long-range dependencies. FasterViT~\cite{hatamizadehFasterViTFastVision2024} adds a learnable ``carrier token" for local-global-local feature refinement.

Quadtree and octree structures are employed for hierarchical attention in unordered 2D and 3D data. Quadtree attention~\cite{tangQuadTreeAttentionVision2022a} dynamically adjusts token granularity in image regions with high attention scores. HST~\cite{heHierarchicalSpatialTransformer2023} uses quadtree-based self-attention in 2D spatial data, while OcTr~\cite{zhouOcTrOctreeBasedTransformer2023} extends this idea to octrees in 3D point clouds, refining high-attention regions. However, these methods suffer from parallelisation challenges due to variable point numbers.

Our~\HOTFormerLoc~overcomes these issues by introducing hierarchical attention into efficient octree-based transformers, enabling global information propagation.

\section{Methodology}
\label{sec:method}
\HOTFormerLoc~uses a Hierarchical Octree Attention mechanism to efficiently exchange global context between local features at multiple scales (\cref{fig:hotformer_overview}). A series of HOTFormer blocks~(\cref{sec:hierarchical_octree_attention}) iteratively process a set of hierarchical features with global and local refinement steps. A pyramid attentional pooling layer (\cref{sec:pyramid_attn_pool}) adaptively fuses this rich set of multi-scale features into a single global descriptor, suitable for LPR across a wide range of sensor configurations and environments as we demonstrate in \cref{sec:experiments}.

\subsection{LPR Problem Formulation}
Let $\mathcal{P}_q = \{ \mathcal{P}_q^{[i]} \in \mathbb{R}^{N_i\times3} \}_{i=1}^{M_q}$ be a set of $M_q$ query submaps, each comprised of a variable number of points $N_i$ captured by a lidar sensor. Let $\mathcal{M}$ be a prior lidar map, which we split into a set of $M_d$ submaps to form the database $ \mathcal{P}_d = \{ \mathcal{P}_d^{[j]} \in \mathbb{R}^{N_j \times 3} \}^{M_d}_{j=1}$. 

Similar to retrieval tasks, in LPR the goal is to retrieve any $\mathcal{P}_d^{[j]}$ captured from the same location as $\mathcal{P}_q^{[i]}$. To this end, our network learns a function $f_\theta : \mathcal{P}_*^{[i]} \rightarrow d_{\mathcal{G}} \in \mathbb{R}^C$ that maps a given lidar submap to a $C$-dimensional global descriptor $d_{\mathcal{G}}$, parameterised by $\theta$. For $\mathcal{P}_d^{[j]}, \mathcal{P}_d^{[k]} \in \mathcal{P}_d $, if $\mathcal{P}_q^{[i]}$ is structurally similar to $\mathcal{P}_d^{[j]}$, but dissimilar to $\mathcal{P}_d^{[k]}$, then we expect $f_\theta(.)$ to satisfy the inequality: 
\vspace{-2mm}
\begin{equation}
    ||f_\theta(\mathcal{P}_q^{[i]}) - f_\theta(\mathcal{P}_d^{[j]})||_2
    < ||f_\theta(\mathcal{P}_q^{[i]}) - f_\theta(\mathcal{P}_d^{[k]})||_2,
    \label{eq:LPR_inequality}
    \vspace{-2mm}
\end{equation}
where $ ||.||_2 $ denotes the $L_2$ distance in feature space.

In most existing LPR tasks, data is collected from a single ground-based platform~\cite{maddernYear1000Km2017,kimMulRanMultimodalRange2020,geigerAreWeReady2012,knightsWildPlacesLargeScaleDataset2023}. Thus, any query $ \mathcal{P}_q^{[i]} $ and structurally-similar submap $ \mathcal{P}_d^{[j]} $ typically have similar distributions (ignoring occlusions and environmental changes). However, when considering data captured by varying lidar configurations from different viewpoints (see \cref{fig:cswildplaces_overview}), $ f_\theta $ must be invariant to the distribution and noise characteristics of each source for~\cref{eq:LPR_inequality} to hold. This proves to be a unique and under-explored challenge, and experiments on our \CSWildPlaces~dataset in \cref{sec:experiments_cswildplaces} show that existing SOTA methods struggle in this setting.

\subsection{Hierarchical Octree Transformer}
\label{sec:hierarchical_octree_transformer}
\noindent\textbf{Octree-based Attention:}
\label{sec:octree_attention}
Motivated by recent advances in 3D Transformers~\cite{liuFlatFormerFlattenedWindow2023,wangOctFormerOctreebasedTransformers2023,wuPointTransformerV32024}, we introduce serialisation-based attention to LPR, addressing the unstructured nature of point clouds which are typically too large for the $\mathcal{O}(N^2 C)$ complexity of self-attention without significant downsampling. In natural environments, the point cloud size of 4096 points common in urban benchmarks offers insufficient detail to represent complex and cluttered scenes. We adopt a sparse octree structure~\cite{meagherOctreeEncodingNew1980} to represent 3D space by recursively subdividing the point cloud into eight equally-sized regions (octants). 
This naturally embeds a spatial hierarchy in the 3D representation, though this is not fully exploited in octree-based self-attention (OSA)~\cite{wangOctFormerOctreebasedTransformers2023}.

In OSA, a function $\phi$ serialises the octree's binary encoding into a Z-order curve, creating a locality-preserving octant sequence~\cite{mortonComputerOrientedGeodetic1966}. This sequence is partitioned into fixed-size local attention windows, reducing self-attention complexity to $\mathcal{O}(k^2 \frac{N}{k} C)$, where $k \ll N$. Compared to 3D transformers with windowed attention~\cite{laiStratifiedTransformer3D2022,sunSWFormerSparseWindow2022,fanEmbracingSingleStride2022,yangSwin3DPretrainedTransformer2023a}, serialisation-based methods are more scalable for large point clouds~\cite{liuFlatFormerFlattenedWindow2023,wuPointTransformerV32024}. See~\cite{wangOctFormerOctreebasedTransformers2023} for further details.

Although efficient and simple, OSA inherits window attention's restricted receptive field compared to full self-attention, while also
failing to fully utilise the octree hierarchy with attention computed for one octree level at a time. This is a key gap, and we argue that multi-scale feature interactions are essential for LPR~\cite{huiPyramidPointCloud2021,guanCrossLoc3DAerialGroundCrossSource2023} where distinctive scene-level descriptors are vital. %

\noindent\textbf{Cylindrical Octree Attention:}
\label{sec:cylindrical_octrees}
Given that raw lidar scans suffer from increased sparsity with distance from the sensor, we propose a simple yet effective modification to the octree structure, particularly suited to the circular pattern of spinning lidar. Typically, octrees are constructed in Cartesian coordinates, where the $(x,y,z)$ dimensions are subdivided to form octants. Instead, we construct octrees in cylindrical coordinates~\cite{sridharaCylindricalCoordinatesLidar2021}, \ie $(\rho,\theta,z)$, to better reflect the distribution of lidar point clouds captured from the ground.

This has two effects. First, octree subdivision now operates radially, causing octants to increase in size (and decrease in resolution) with distance from the sensor, resulting in higher resolutions for octants near the sensor where point density is typically highest. Second, the octree serialisation function $\phi$ now operates cylindrically, changing the octant ordering into cylindrical local attention windows. %

\begin{figure}[t]
  \centering
  \begin{subfigure}{0.47\linewidth}
    \includegraphics[width=.99\linewidth]{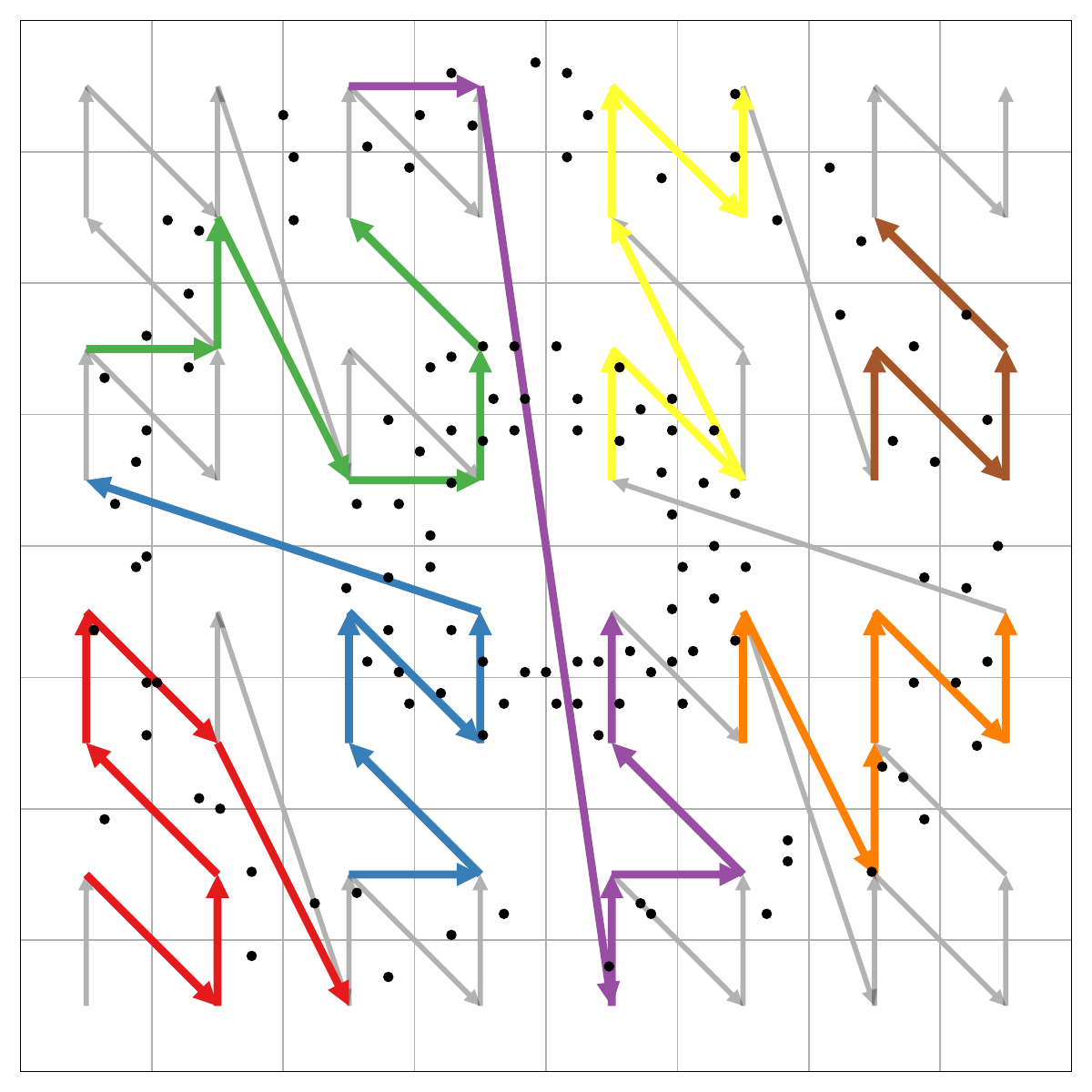}
    \caption{\small{Cartesian octree}}
    \label{fig:cart_S_windows}
  \end{subfigure}
  \hfill
  \begin{subfigure}{0.47\linewidth}
    \includegraphics[width=.99\linewidth]{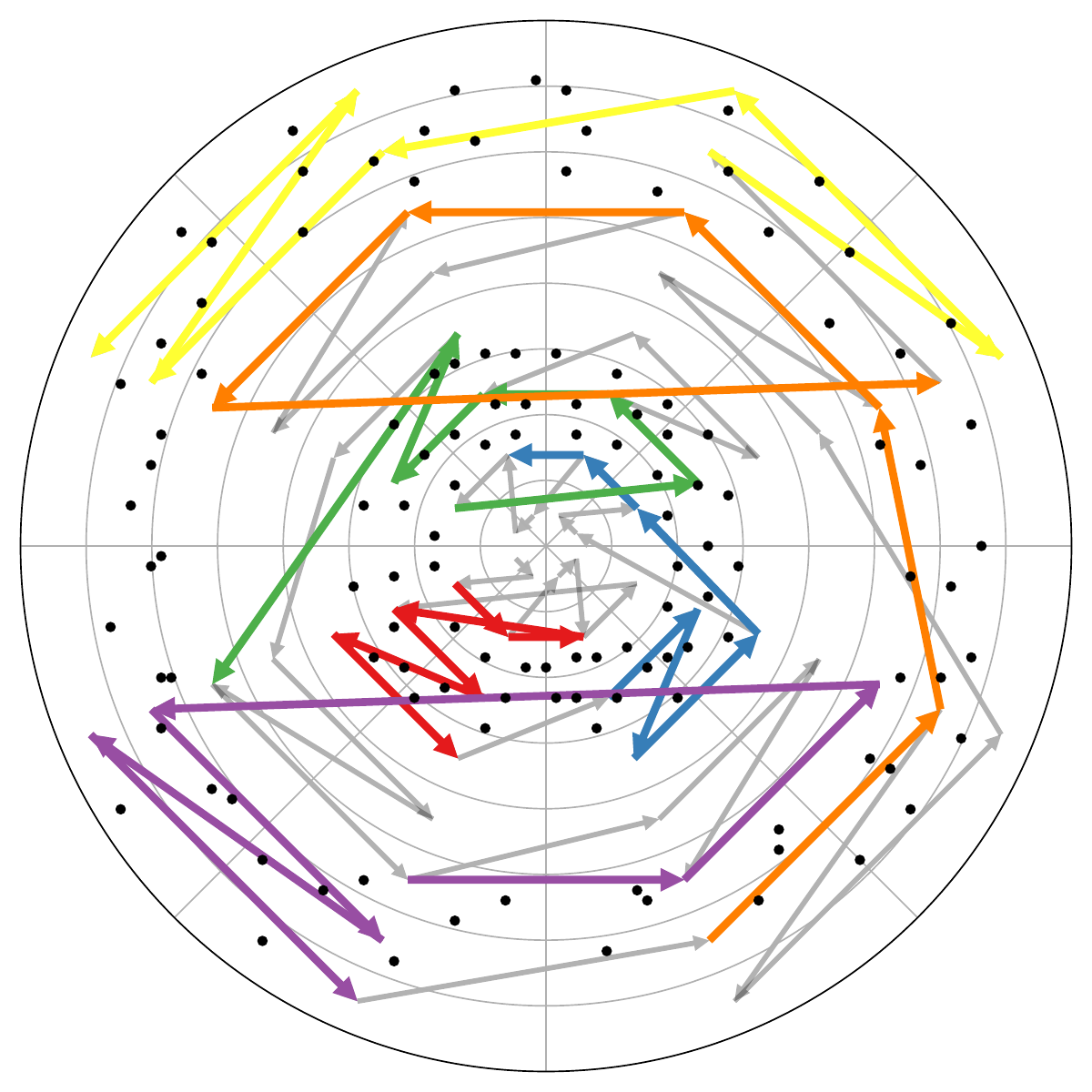}
    \caption{\small{Cylindrical octree}}
    \label{fig:polar_S_windows}
  \end{subfigure}
  \vspace{-2mm}
  \caption{\small{Cartesian VS cylindrical attention window serialisation (each window indicated by the arrow colour) for the 2D equivalent of an octree with depth $d=3$ and window size $k=7$. 
  }}
  \label{fig:zorder_cart_polar_comparison}
  \vspace{-6mm}
\end{figure}

The effect of this on the 2D-equivalent of an octree is seen in \cref{fig:zorder_cart_polar_comparison}, which visualises attention windows for two concentric rings of points with differing densities (mimicking the behaviour of spinning lidar in outdoor scenes).
Cartesian octree attention windows cover a uniform area and ignore the underlying point density, whereas the cylindrical octree attention windows respect the point distribution, with fine-grained windows near the sensor, and sparser windows at lower-density regions further from away. We demonstrate the effectiveness of this approach on point clouds captured in natural environments in \cref{tab:ablation_cylindrical}. See \cref{fig:supp_cylindrical_attention_hierarchy} in the Supplementary for further detail.

\noindent\textbf{Hierarchical Octree Attention:}
\label{sec:hierarchical_octree_attention}
Inspired by image-based hierarchical window attention approaches~\cite{yangFocalAttentionLongRange2021a,chuTwinsRevisitingDesign2021a,panEdgeViTsCompetingLightweight2022,tangQuadTreeAttentionVision2022a,hatamizadehFasterViTFastVision2024}, we propose a novel hierarchical octree transformer (HOTFormer) block to unlock the potential of octree-based attention for multi-scale representation learning from point clouds. We introduce \textit{\carriertoken s} to capture global feature interactions between multiple scales of a hierarchical octree feature pyramid, and adopt a two-step process to iteratively propagate contextual information and refine features in a global-to-local fashion. \cref{fig:hotformer_overview} illustrates our approach.

To initialise our feature pyramid, an input point cloud $P \in \mathbb{R}^{N \times 3}$ with $N$ points is encoded into a sparse octree of depth $d$, where unoccupied octants are pruned from subsequent levels during construction. Embeddings are generated with a sparse octree-based convolution~\cite{wangOCNNOctreebasedConvolutional2017a} stem followed by several OSA transformer blocks to produce an initial local feature map $F_0 \in \mathbb{R}^{N_{d} \times C}$, where $N_d$ is the number of non-empty octants at initial octree depth $d$
and $C$ is the feature dimension. Starting from $F_0$, we initialise the hierarchical octree feature pyramid with:
\vspace{-2mm}
\begin{equation}
    F = \left\{ \mathrm{LN}(\mathrm{DS}(F_{l-1})) \in \mathbb{R}^{N_{d-l} \times C} \right\}_{l=1}^{L},
    \vspace{-2mm}
\end{equation}
where $\mathrm{LN}$ is layer normalisation~\cite{leibaLayerNormalization2016}, $\mathrm{DS}$ is a downsampling layer composed of a sparse convolution layer with kernel size and stride of 2, and $F$ is the set of feature maps $F_l$ across all levels $l=1,...,L$ of the pyramid. 

Each $ F_l $ is generated by downsampling and normalising the previous level's feature map, $ F_{l-1} $, resulting in progressively coarser $F_l$'s as the octree is traversed toward the root. Consequently, each $ F_l $ will have $ N_{d-l} $ local tokens capturing features with increasing spatial coverage at higher levels. Ideally, self-attention across local features from all levels would capture a multi-scale representation. However, self-attention's quadratic complexity makes this prohibitive. Addressing this bottleneck, we propose \textit{\carriertoken s} to efficiently \textit{relay} contextual cues between distant regions.

Conceptually, \carriertoken s act as proxies that distill key information from local regions at different octree granularities into compact representations. Within each HOTFormer block (see \cref{fig:hierarchical_attention}), this representation is used to model long-range feature interactions through relay token self-attention (RTSA) layers, whilst being tractable in large-scale point clouds. A hierarchical octree self-attention (H-OSA) layer then computes window attention between refined \carriertoken s and corresponding local features to efficiently propagate global context to local feature maps. Furthermore, by allowing \carriertoken s from all pyramid levels to interact during RTSA, we induce \textit{hierarchical} feature interactions within the octree. This enables the network to efficiently capture multi-scale features and handle variations in point density and distributions across diverse sources.

Before the series of HOTFormer blocks, we reshape each $F_l$ into local attention windows of size $k$ with the serialisation function $\phi : F_l \rightarrow \hat{F}_l \in \mathbb{R}^{w_l \times k \times C}$, where $w_l = \frac{N_{d-l}}{k}$ is the number of local attention windows at level $l$. Then, for all local attention windows in $\hat{F}_l$, we introduce a set of \carriertoken s $\mathrm{RT}_l$ which summarise each attention window. Formally, we initialise \carriertoken s at each pyramid level:
\vspace{-2mm}
\begin{equation}
    \mathrm{RT}_l = \mathrm{AvgPool}_{w_l \times k \times C \rightarrow w_l \times C}(\hat{F}_l) + \mathrm{ADaPE}(\Psi_l),
    \label{eq:rt_init}
    \vspace{-2mm}
\end{equation}
where $\mathrm{AvgPool}$ pools the $k$ local tokens in each window, and $\mathrm{ADaPE}$ is a novel absolute distribution-aware positional encoding, described in the following section.

The \carriertoken s $\mathrm{RT}_l$ and local windows $\hat{F}_l$ are processed by $M$ HOTFormer blocks, composed of successive RTSA and H-OSA layers. In each RTSA layer, \carriertoken s from each level $l$ are concatenated with $\mathrm{RT}' = \mathrm{Concat}(\{\mathrm{RT}_l\}_{l=1}^L,\ \mathrm{dim}=0)$, where $\mathrm{RT}' \in \mathbb{R}^{w_{\mathrm{total}} \times C}$.
The multi-scale \carriertoken s are processed by a transformer:
\vspace{-6mm}
\begin{equation}
\label{eq:begin_hotf_block}
\resizebox{0.9\hsize}{!}{$
    \mathrm{RT}' = \mathrm{RT}' + \mathrm{MHSA}(\mathrm{LN}(\mathrm{RT}')),~
    \mathrm{RT}' = \mathrm{RT}' + \mathrm{FFN}(\mathrm{LN}(\mathrm{RT}')), 
$}
\vspace{-2mm}
\end{equation}
where $\mathrm{MHSA}$ denotes multi-head self-attention~\cite{vaswaniAttentionAllYou2017b}, and $\mathrm{FFN}$ is a 2-layer MLP with GeLU~\cite{hendrycksBridgingNonlinearitiesStochastic2016} activation. The \carriertoken s are then split back to their respective pyramid levels with $\mathrm{RT}_1, ..., \mathrm{RT}_L = \mathrm{Split}(\mathrm{RT}')$.

Next, the interaction between \carriertoken s and local tokens is computed using a H-OSA layer at each pyramid level. We first apply a conditional positional encoding (CPE)~\cite{chuConditionalPositionalEncodings2021} to local tokens with $\hat{F_l} = \hat{F_l} + \mathrm{CPE}(\hat{F_l})$, implemented with an octree-based depth-wise convolution layer. Local attention windows are concatenated with their corresponding \carriertoken\ to create \textit{hierarchical attention windows} for each level $l$ with $\tilde{F_l} = \mathrm{Concat}(\hat{F}_{l}, \mathrm{RT}_{l},\ \mathrm{dim}=1)$, where $\tilde{F_l} \in \mathbb{R}^{w_l \times (1+k) \times C}$. All levels are then processed individually with another set of transformer blocks:
\vspace{-2mm}
\begin{equation}
\resizebox{0.9\hsize}{!}{$
    \tilde{F_l} = \tilde{F_l} + \mathrm{MHSA}(\mathrm{LN}(\tilde{F_l})),~
    \tilde{F_l} = \tilde{F_l} + \mathrm{FFN}(\mathrm{LN}(\tilde{F_l})).
$}
\vspace{-2mm}
\end{equation}

Local windows and \carriertoken s from each level are separated with $\hat{F}_{l}, \mathrm{RT}_{l} = \mathrm{Split}(\tilde{F_l})$, ready to be processed in subsequent HOTFormer blocks.
This alternating process of global-local attention is repeated $M$ times, and the resultant multi-scale local attention windows $\hat{F_l}$ return to feature maps with the inverse serialisation function $\phi^{-1}: \hat{F_l} \rightarrow F_l \in \mathbb{R}^{N_{d-l} \times C}$. The refined octree feature pyramid $F$ is sent to a pyramid attentional pooling layer for aggregation into a single global descriptor. We provide complexity analysis of these novel layers and visualisations of the learned multi-scale attention patterns in Supplementary Materials.

\noindent
\textbf{Distribution-aware Positional Encoding:}
\label{sec:adape}
One challenge posed by \carriertoken s is defining an appropriate positional encoding. Octree attention windows and their corresponding \carriertoken s represent a sparse, non-uniform region compared to windows partitioned on a regular grid. As such, convolution-based CPE is not directly computable, and a relative position encoding is difficult to define. The positional relationship between multi-scale \carriertoken s must also be considered, as \carriertoken s in coarser levels represent larger regions than those in fine-grained levels.

To address this, we propose an Absolute Distribution-aware Positional Encoding (ADaPE) by injecting knowledge of the underlying point distribution into an absolute positional encoding. For a given relay token $\mathrm{RT}_l^{[i]}$ and corresponding attention window $\Hat{F}_l^{[i]}$, we compute the centroid $\mu$ and sample covariance matrix $\Sigma$ of point coordinates in the window. We then construct a tuple $\Psi_l^{[i]}$ with $\mu$ and the flattened upper triangular of $\Sigma$ such that $\Psi_l^{[i]} = (\mu_x, \mu_y, \mu_z, \sigma_x, \sigma_y, \sigma_z, \sigma_{xy}, \sigma_{xz}, \sigma_{yz})$. This is repeated for \carriertoken s at all pyramid levels and each $\Psi_l$ is processed by a shared 2-layer MLP to encode a higher-dimensional representation. We inject this positional encoding directly after relay token initialisation in \cref{eq:rt_init}.
    
This approach unlocks the full potential of RTSA, as the network can learn a representation that aligns the attention scores of adjacent \carriertoken s, whilst being aware when one \carriertoken~occupies a larger region. We provide ablations demonstrating the effectiveness of ADaPE in \cref{tab:ablation_components}.

\subsection{Pyramid Attentional Pooling}
\label{sec:pyramid_attn_pool}
To best utilise the set of hierarchical octree features, we propose a novel pyramid attentional pooling mechanism to aggregate multi-resolution tokens into a distinctive global descriptor $d_{\mathcal{G}} \in \mathbb{R}^C$ whilst adaptively filtering out irrelevant tokens. Our motivation arises from the observation that there is rarely a single best spatial feature resolution to represent point clouds captured from various environments or sources, hence we consider a range of resolutions during pooling to improve the generality of HOTFormerLoc.

Attention pooling~\cite{goswamiSALSASwiftAdaptive2024} employs a learnable query matrix to pool a variable number of tokens into a fixed number of tokens. These queries provide flexibility for the network to learn distinctive clusters of tokens that best represent the environment, and we introduce a set of learnable queries $Q^{\theta}_l \in \mathbb{R}^{q_l \times C}$ for each octree feature pyramid level. Unlike previous approaches~\cite{pengAttentionalPyramidPooling2021}, our approach can handle the variable number of local tokens from each level whilst retaining linear computational complexity. Reflecting the pyramidal nature of local features, we pool more tokens from fine-grained pyramid levels, and fewer tokens from coarser levels. Pyramid attentional pooling can be formulated as:
\vspace{-2mm}
\begin{equation}
\resizebox{0.9\hsize}{!}{$
     \Omega_l =  \mathrm{softmax} \left( \frac{Q_l^{\theta} F_l^T}{\sqrt{C}} \right) F_l,~
     \Omega' = \mathrm{Concat}(\{\Omega_l\}_{l=1}^{L},\ \mathrm{dim}=0),
$}
    \vspace{-2mm}
\end{equation}
where $\Omega_l \in \mathbb{R}^{q_l \times C}$ is the set of pooled tokens from pyramid level $l$, concatenated to form $\Omega' \in \mathbb{R}^{q_{\mathrm{total}} \times C}$, and the tokens $F_l$ are used as the key and value matrices in attention.

We enhance interactions between pooled multi-scale tokens $\Omega'$ with a token fuser~\cite{ali-beyMixVPRFeatureMixing2023}, composed of four 2-layer MLPs with layer normalisation~\cite{leibaLayerNormalization2016} and GeLU~\cite{hendrycksBridgingNonlinearitiesStochastic2016} activations. The fused tokens pass through an MLP-Mixer~\cite{tolstikhinMLPMixerAllMLPArchitecture2021}, where a series of token-mixing and channel-mixing MLPs reduce the number and dimensionality of pooled tokens, which are flattened and $L_2$-normalised to produce the global descriptor $d_\mathcal{G} \in \mathbb{R}^C$. We provide further analysis of pyramid attentional pooling in the Supplementary Material.

\begin{figure}[t]
  \centering
   \includegraphics[width=0.98\linewidth, trim={0 0 0 0.8cm},clip]{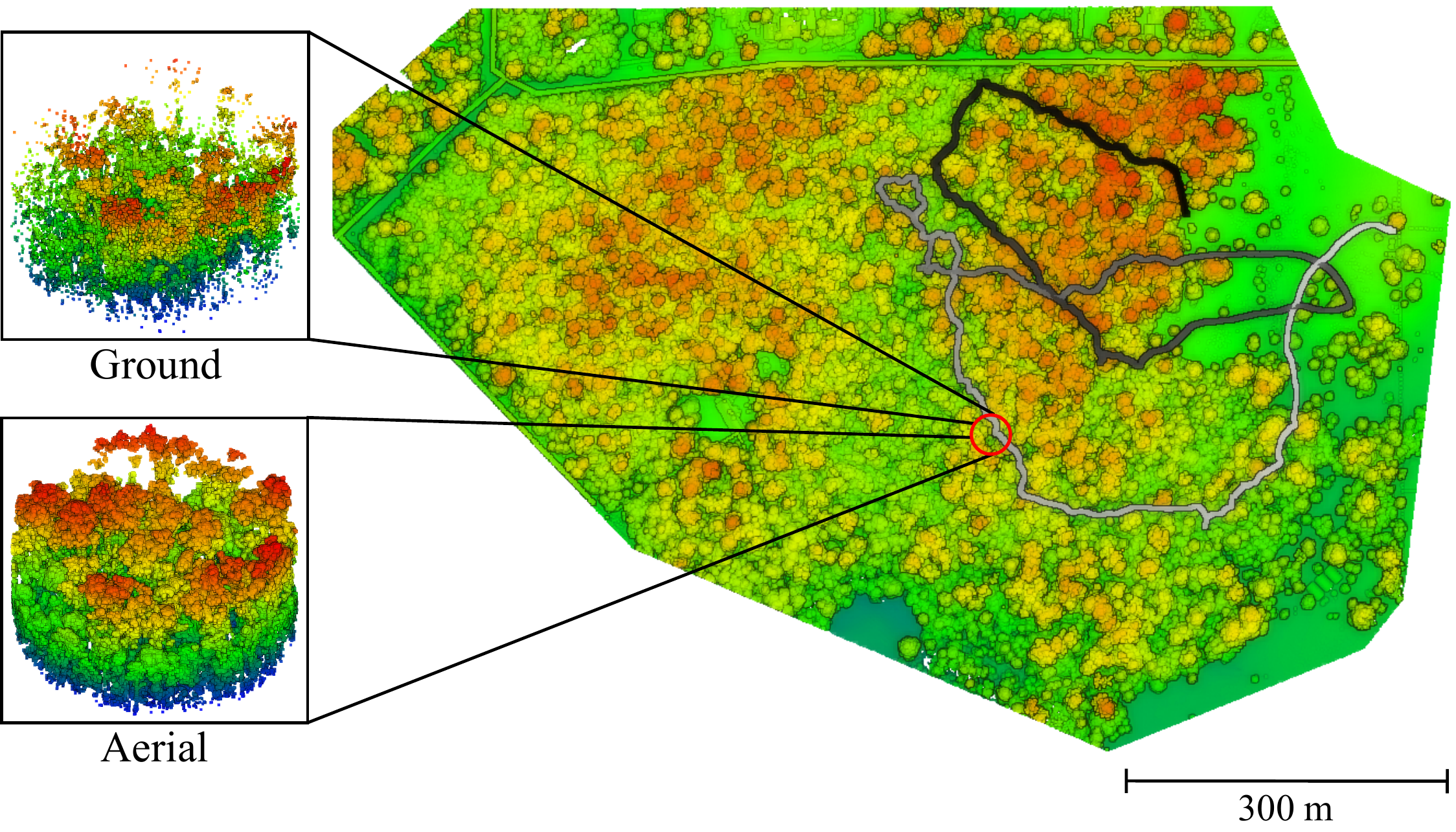}
   \vspace{-2mm}
   \caption{\small{Bird's-eye view of an aerial map from \CSWildPlaces, colourised by height, with the ground sequence trajectory overlaid. %
   Ground-aerial submap differences pose major challenges.}}
   \label{fig:cswildplaces_overview}
   \vspace{-5mm}
\end{figure}

\section{\CSWildPlaces~Dataset}
\vspace{-1mm}
\label{sec:dataset}

We present the first benchmark for cross-source LPR in unstructured, natural environments, dubbed \CSWildPlaces, featuring high-resolution ground and aerial lidar submaps collected over three years from four forests in Brisbane, Australia. We build upon the 8 ground sequences introduced in the Wild-Places dataset~\cite{knightsWildPlacesLargeScaleDataset2023} by capturing aerial lidar scans of Karawatha and Venman forests, forming our ``Baseline" set. We further introduce ground and aerial lidar scans from two new forests: QCAT and Samford (\cref{fig:cswildplaces_overview}), forming our ``Unseen" testing set. \cref{tab:dataset_comparison} compares our \CSWildPlaces~dataset with popular LPR benchmarks.

\noindent
\textbf{Data Collection}:
Ground data collection uses a handheld perception pack with spinning VLP-16 lidar sensor. We capture 2 sequences on foot in QCAT and Samford, supplementing the 8 Wild-Places sequences for a total of $36.4~km$ traversal over $13.8~km$ of trails. To generate globally consistent maps and near-ground truth trajectories, we use Wildcat SLAM~\cite{ramezaniWildcatOnlineContinuousTime2022}, integrating GPS, IMU, and lidar.

For aerial data collection, we deployed two drone configurations. For Karawatha, Venman, and QCAT, we used a DJI M300 quadcopter with a VLP-32C lidar sensor. For Samford, we used an Acecore NOA hexacopter equipped with a RIEGL VUX-120 pushbroom lidar. Both drones flew in a lawnmower pattern over forested areas at a consistent height of $\sim\!50$ -- $100~m$ above the canopy.
GPS RTK is used for all aerial scans to ensure precise geo-registration in UTM coordinates. We align overlapping ground and aerial areas using iterative closest point~\cite{besl1992method} until the RMSE between correspondences is $\leq 0.5~m$. See Supplementary Materials for further details and environment visualisations.

\noindent
\textbf{Submap Generation:}
We follow two protocols to generate lidar submaps suitable for LPR. Ground submaps are sampled at $0.5~Hz$ along each trajectory, aggregating all points captured within a one second sliding window of the corresponding timestamp, within a $30~m$ horizontal radius. Points are stored in the submap’s local coordinates, along with the 6-DoF pose in UTM coordinates.

Aerial submaps are uniformly sampled from a $10~m$ spaced grid spanning the aerial map. To create a realistic scenario, grid borders are set to sample a much larger area than is covered by the ground traversals. For consistency with ground submaps, we limit submaps to a $30~m$ horizontal radius. This produces a set of overlapping aerial patches that form a comprehensive database covering each forest. 

We further post-process submaps, removing all points situated on the ground plane using a Cloth Simulation Filter (CSF)~\cite{zhangEasytoUseAirborneLiDAR2016}. To save computation, we voxel downsample submaps with voxel size of $0.8~m$, generating submaps with an average of $28K$ points.

\noindent
\textbf{Training and Testing Splits:}
We train LPR methods using submaps from the Baseline set, and withhold a disjoint set of submaps for evaluation, following the test regions of Wild-Places. To prevent information leakage between training and evaluation, we exclude any submaps from training that overlap the evaluation queries. For optimising triplet-based losses, we construct training tuples with a $15~m$ positive threshold, and $60~m$ negative threshold.

During evaluation, we use the withheld Baseline ground submaps as queries, and all aerial submaps as a per-forest database.
We test generalisation to new environments on our Unseen test set, using all submaps in the set to form the ground queries and per-forest aerial database.
We consider a true positive retrieval threshold of $30~m$ during evaluation.

\begin{table}[t]
  \centering
  \huge
  \resizebox{\linewidth}{!}{
  \begin{tabular}{@{}l c c c@{}}
    \toprule
     Dataset& \makecell[cc]{Oxford \\ RobotCar~\cite{maddernYear1000Km2017}} & CS-Campus3D~\cite{guanCrossLoc3DAerialGroundCrossSource2023} & \CSWildPlaces~(Ours) \\
    \midrule 
    Environment & Urban (Street) & Urban (Campus) & Forest \\
    Viewpoint & Ground & Ground, Aerial & Ground, Aerial \\
    Platform & Car & Mobile Robot / Airplane & Handheld / UAV \\
    Length / Coverage & 1000~km & 7.8~km / 5.5~km$^2$ & 36.4~km / 3.7~km$^2$ \\
    Avg. Point Resolution & 4096 & 4096 & 28K \\
    \makecell[cl]{Num. Ground Submaps \\ (training / testing)} & 21711 / 3030 & 6167 / 1538 & 43656 / 16037 + 2086 \\
    \makecell[cl]{Num. Aerial Submaps \\ (database)} & N/A & 27520 & 28686 + 4950 \\
    Submap Diameter & 20-25~m & 100~m & 60~m \\
    Retrieval Threshold & 25~m & 100~m & 30~m \\
    \bottomrule
  \end{tabular}
  }
  \vspace{-2mm}
  \caption{\small{Comparison of \CSWildPlaces~with popular LPR benchmarks. For \CSWildPlaces~submaps, $X+Y$ indicates $X$ submaps in the Baseline test set, and $Y$ submaps in the Unseen test set. %
  }}
  \label{tab:dataset_comparison}
  \vspace{-5mm}
\end{table}

\section{Experiments}
\label{sec:experiments}

\noindent
\textbf{Datasets and Evaluation Criteria:}
To demonstrate our method's versatility, we conduct experiments on Oxford RobotCar~\cite{maddernYear1000Km2017}, CS-Campus3D~\cite{guanCrossLoc3DAerialGroundCrossSource2023}, and Wild-Places~\cite{knightsWildPlacesLargeScaleDataset2023}, using the established training and testing splits for each, alongside our \CSWildPlaces~dataset proposed in \cref{sec:dataset}. This selection tests diverse scenarios, such as ground-to-ground and ground-to-aerial LPR in urban and forest environments. 

We report AR@N (including variants like AR@1 and AR@1$\%$), a standard PR performance metric. AR@N quantifies the percentage of correctly localised queries where at least one of the top-N database predictions matches the query. We also report the mean reciprocal rank (MRR) for consistency with Wild-Places, defined as $ \mathrm{MRR} = \frac{1}{M_q} \sum_{i=1}^{M_q} \frac{1}{\mathrm{rank}_i} $, where $\mathrm{rank}_i$ is the ranking of the first true positive retrieval for each query submap.

\begin{figure}
    \centering
    \includegraphics[width=0.95\linewidth]{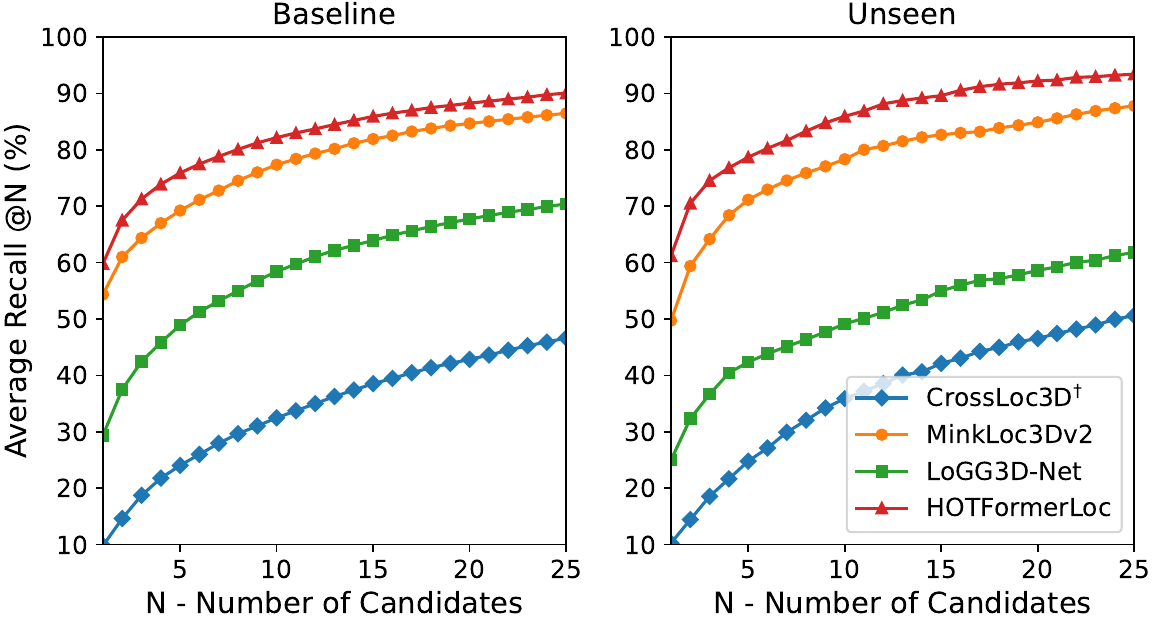}
    \vspace{-2mm}
    \caption{\small{Recall@N curves of SOTA on \CSWildPlaces. CrossLoc3D$^\dagger$ indicates submaps were randomly downsampled to 4096 points for compatibility with the method.}}
    \label{fig:cswildplaces_ground-aerial}
    \vspace{-3mm}
\end{figure}

\noindent
\textbf{Implementation Details:}
Our network uses $M=10$ HOTFormer blocks operating on $L=3$ pyramid levels with channel size of $C=256$, and an initial octree depth of $d=7$ with attention window size of $k=48$ for Oxford and Wild-Places, and $k=64$ for CS-Campus3D and \CSWildPlaces. For pyramid attention pooling, we set the number of pooled tokens per level to $q=[74,36,18]$ and the global descriptor size to 256 for a fair comparison with baselines.

All experiments are performed on a single NVIDIA H100 GPU. We follow the training protocol of~\cite{komorowskiImprovingPointCloud2022}, utilising a Truncated Smooth-Average Precision (TSAP) loss and large batch size of 2048, trained on a single GPU by computing sub-batches with multistaged backpropagation~\cite{revaudLearningAveragePrecision2019}.

We use the Adam~\cite{kingmaAdamMethodStochastic2014} optimiser with weight decay of $1e^{-4}$ and learning rate (LR) in the range [$5e^{-4}$, $3e^{-3}$] depending on the dataset. We adopt data augmentations including random flips, random rotations of $\pm 180^\circ$, random translation, random point jitter, and random block removal. Further, we use a Memory-Efficient Sharpness-Aware~\cite{duSharpnessAwareTrainingFree2022} auxiliary loss, enabled after $15\%$ of training epochs, to aid generalisation by encouraging convergence to a flat minima.

\begin{table}[t]
  \centering
  \resizebox{0.35\textwidth}{!}{
  \begin{tabular}{@{}lcccccccc@{}}
    \toprule
    &&&&&&& \multicolumn{2}{c}{CS-Campus3D} \\
    Method &&&&&&& AR@1 $\uparrow$ & AR@1\% $\uparrow$ \\
    \midrule
    PointNetVLAD~\cite{uyPointNetVLADDeepPoint2018} &&&&&&& 19.1 & 43.6 \\
    TransLoc3D~\cite{xuTransLoc3DPointCloud2022} &&&&&&& 43.0 & 80.6 \\ 
    MinkLoc3Dv2~\cite{komorowskiImprovingPointCloud2022} &&&&&&& 52.5 & 83.5 \\
    CrossLoc3D~\cite{guanCrossLoc3DAerialGroundCrossSource2023} &&&&&&& \underline{70.7} & \underline{85.7} \\
    \textbf{\HOTFormerLoc~(Ours)} &&&&&&& \textbf{80.4} & \textbf{94.9} \\
    \bottomrule
  \end{tabular}
  }
  \vspace{-2mm}
  \caption{\small{Comparison of SOTA on CS-Campus3D~\cite{guanCrossLoc3DAerialGroundCrossSource2023} with ground-only queries, and aerial-only database.}}
  \label{tab:campus3d_g/g+a_results}
  \vspace{-5mm}
\end{table}

\subsection{Comparison with SOTA}
\label{sec:experiments_cswildplaces}
\textbf{\CSWildPlaces:}~In \cref{fig:cswildplaces_ground-aerial} we demonstrate the performance of the proposed HOTFormerLoc on our \CSWildPlaces~dataset, trained for 100 epochs with a LR of $8e^{-4}$, reduced by a factor of 10 after 50 epochs. On the Baseline and Unseen evaluation sets, HOTFormerLoc achieves an improvement in AR@1 of $5.5\%$ -- $11.5\%$, and an improvement in AR@1\% of $3.6\%$ -- $4.5\%$, respectively.
As CrossLoc3D~\cite{guanCrossLoc3DAerialGroundCrossSource2023} requires input point clouds to have exactly 4096 points, we provide results for this method by training on a variant of \CSWildPlaces~with submaps randomly downsampled to 4096 points. CrossLoc3D's mean AR@1 of 10.1\% across both evaluation sets clearly shows the limitations of this approach. 

While LoGG3D-Net~\cite{vidanapathiranaLoGG3DNetLocallyGuided2022} is the top-performing method on Wild-Places~\cite{knightsWildPlacesLargeScaleDataset2023}, we see its performance drop considerably on our dataset using the same configuration. We hypothesise that the local-consistency loss introduced in LoGG3D-Net is ill-suited to cross-source data, as the lower overlap between submaps leads to fewer point correspondences for optimisation. 

\begin{table}[t]
  \centering
  \resizebox{\linewidth}{!}{
  \begin{tabular}{@{}lcc cc|cc@{}}
    \toprule
    \Huge
    & \multicolumn{2}{c}{Karawatha} & \multicolumn{2}{c}{Venman} & \multicolumn{2}{c}{Mean} \\
    Method & AR@1 $\uparrow$ & MRR $\uparrow$ & AR@1 $\uparrow$ & MRR $\uparrow$ & AR@1 $\uparrow$ & MRR $\uparrow$ \\
    \midrule
    TransLoc3D~\cite{xuTransLoc3DPointCloud2022} & 46.1 & 50.2 & 50.2 & 66.2 & 48.2 & 58.2 \\ 
    MinkLoc3Dv2~\cite{komorowskiImprovingPointCloud2022} & 67.8 & 79.2 & 75.8 & 84.9 & 71.8 & 82.0 \\
    LoGG3D-Net~\cite{vidanapathiranaLoGG3DNetLocallyGuided2022} & \textbf{74.7} & \textbf{83.7} & \underline{79.8} & \underline{87.3} & \textbf{77.3} & \textbf{85.5} \\
    LoGG3D-Net$^1$~\cite{vidanapathiranaLoGG3DNetLocallyGuided2022} & 57.9 & 72.4 & 63.0 & 75.5 & 60.5 & 73.9 \\
    \textbf{\HOTFormerLoc$^\dagger$ (Ours)} & \underline{69.6} & \underline{80.1} & \textbf{80.1} & \textbf{87.4} & \underline{74.8} & \underline{83.7} \\
    \bottomrule
  \end{tabular}
  }
  \vspace{-2mm}
  \caption{\small{Comparison on Wild-Places~\cite{knightsWildPlacesLargeScaleDataset2023}. \HOTFormerLoc$^\dagger$ denotes cylindrical octree windows. LoGG3D-Net$^1$ indicates training the network using a 256-dimensional global descriptor, as opposed to the 1024-dimensional descriptor reported in Wild-Places.}}
  \label{tab:wildplaces_interseq_results}
  \vspace{-5mm}
\end{table}

\begin{table*}[t]
  \centering
  \resizebox{0.90\linewidth}{!}{
  \begin{tabular}{lcc cc cc cc|cc}
    \toprule
    & \multicolumn{2}{c}{Oxford} & \multicolumn{2}{c}{U.S} & \multicolumn{2}{c}{R.A.} & \multicolumn{2}{c}{B.D.} & \multicolumn{2}{c}{Mean} \\
    Method & AR@1 $\uparrow$ & AR@1\% $\uparrow$ & AR@1 $\uparrow$ & AR@1\% $\uparrow$ & AR@1 $\uparrow$ & AR@1\% $\uparrow$ & AR@1 $\uparrow$ & AR@1\% $\uparrow$ & AR@1 $\uparrow$ & AR@1\% $\uparrow$ \\
    \midrule
    PointNetVLAD~\cite{uyPointNetVLADDeepPoint2018} & 62.8 & 80.3 & 63.2 & 72.6 & 56.1 & 60.3 & 57.2 & 65.3 & 59.8 & 69.6 \\
    PPT-Net~\cite{huiPyramidPointCloud2021} & 93.5 & 98.1 & 90.1 & \underline{97.5} & 84.1 & 93.3 & \underline{84.6} & 90.0 & 88.1 & 94.7 \\
    TransLoc3D~\cite{xuTransLoc3DPointCloud2022} & 95.0 & 98.5 & --- & 94.9 & --- & 91.5 & --- & 88.4 & --- & 93.3 \\
    MinkLoc3Dv2~\cite{komorowskiImprovingPointCloud2022} & \underline{96.3} & \textbf{98.9} & \underline{90.9} & 96.7 & \underline{86.5} & \underline{93.8} & \underline{86.3} & \underline{91.2} & \underline{90.0} & \underline{95.1} \\
    CrossLoc3D~\cite{guanCrossLoc3DAerialGroundCrossSource2023} & 94.4 & 98.6 & 82.5 & 93.2 & 78.9 & 88.6 & 80.5 & 87.0 & 84.1 & 91.9 \\    
    \textbf{\HOTFormerLoc~(Ours)} & \textbf{96.4} & \underline{98.8} & \textbf{92.3} & \textbf{97.9} & \textbf{89.2} & \textbf{94.8} & \textbf{90.4} & \textbf{94.4} & \textbf{92.1} & \textbf{96.5} \\ %
    \bottomrule
  \end{tabular}
  }
  \vspace{-2mm}
  \caption{\small{Comparison of SOTA on Oxford RobotCar~\cite{maddernYear1000Km2017} using the baseline evaluation setting and dataset introduced by~\cite{uyPointNetVLADDeepPoint2018}.}}
  \label{tab:oxford_results}
  \vspace{-5mm}
\end{table*}

\noindent
\textbf{CS-Campus3D:}
In \cref{tab:campus3d_g/g+a_results}, we present the evaluation results on CS-Campus3D, training our method for 300 epochs with a LR of $5e^{-4}$, reduced by a factor of 10 after 250 epochs. Our approach shows an improvement of $9.7\%$ and $9.2\%$ in AR@1 and AR@1\%, respectively. Most notably, we exceed the performance of CrossLoc3D~\cite{guanCrossLoc3DAerialGroundCrossSource2023}, which employs a diffusion-inspired refinement step to specifically address the cross-source challenge. This highlights the versatility of our hierarchical attention approach, which is capable of learning general representations to achieve SOTA performance in both single- and cross-source settings.

\begin{table}
  \centering
  \resizebox{0.85\linewidth}{!}{
  \begin{tabular}{lcc}
    \toprule
    Method & Parameters (M) & Inference Time (ms) \\
    \midrule
    MinkLoc3Dv2~\cite{komorowskiImprovingPointCloud2022} & 2.6 & 103.2 \\
    LoGG3D-Net~\cite{vidanapathiranaLoGG3DNetLocallyGuided2022} & 8.8 & 209.8 \\
    \textbf{\HOTFormerLoc~(Ours)} & 35.4 & 270.0 \\
    \bottomrule
  \end{tabular}
  }
  \vspace{-2mm}
  \caption{\small{Efficiency comparison with SOTA LPR methods using submaps from \CSWildPlaces~with $\sim$28K points.} 
  }
  \label{tab:effiency_comparison}
  \vspace{-3mm}
\end{table}

\begin{table}
  \centering
  \resizebox{0.9\linewidth}{!}{
  \begin{tabular}{l ccc}
    \toprule
     & Oxford & CS-Campus3D & \CSWildPlaces \\
    Ablation & AR@1 (Mean) & AR@1 & AR@1 (Mean) \\
    \midrule
    Relay Tokens Disabled & -2.5\% & -4.5\% & -4.7\% \\
    ADaPE Disabled & -1.0\% & -2.9\% & -1.8\% \\
    L=2 Pyramid Levels & -3.1\% & -5.1\% & -3.8\% \\
    GeM Pooling & -7.3\% & -22.5\% & -15.1\% \\
    Pyramid GeM Pooling & -2.8\% & -3.4\% & -12.3\% \\
    \bottomrule
  \end{tabular}
  }
  \vspace{-2mm}
  \caption{\small{Ablation study on the effectiveness of \HOTFormerLoc~components on Oxford, CS-Campus3D and \CSWildPlaces.}}
  \label{tab:ablation_components}
  \vspace{-5mm}
\end{table}

\noindent
\textbf{Wild-Places:}~In~\cref{tab:wildplaces_interseq_results}, we report evaluation results on Wild-Places under the inter-sequence evaluation setting, training our method for 100 epochs with a LR of $3e^{-3}$, reduced by a factor of 10 after 30 epochs. LoGG3D-Net~\cite{vidanapathiranaLoGG3DNetLocallyGuided2022} remains the highest performing method by a margin of 2.5\% and 1.8\% in AR@1 and MRR, respectively, but we achieve a gain of $5.5\%$ and $3.5\%$ in AR@1 and MRR over MinkLoc3Dv2. However, we note that LoGG3D-Net is trained on Wild-Places with a global descriptor size of 1024, compared to our compact descriptor of size 256. 

\noindent
\textbf{Oxford:}
\cref{tab:oxford_results} reports evaluation results on Oxford RobotCar using the baseline evaluation setting and dataset introduced by~\cite{uyPointNetVLADDeepPoint2018}, training our method for 150 epochs with a LR of $5e^{-4}$, reduced by a factor of 10 after 100 epochs. We outperform previous SOTA methods, showing improved generalisation on the unseen R.A. and B.D. environments with an increase of $2.7\%$ and $4.1\%$ in AR@1, respectively. %

\noindent
\textbf{Runtime Analysis:}
\label{sec:runtime_analysis}
In \cref{tab:effiency_comparison}, we provide comparisons of parameter count and runtime for SOTA LPR methods capable of handling large point clouds, on a machine equipped with a NVIDIA A3000 mobile GPU and 12-core Intel Xeon W-11855M CPU. Naturally, HOTFormerLoc is bulkier than previous approaches due to the use of transformers over sparse CNNs. However, our approach is still fast enough for deployment online, 
and requires only 1.2GB GPU memory during inference, suitable for edge devices.

\subsection{Ablation Study}
\textbf{HOTFormerLoc Components:}
In \cref{tab:ablation_components}, we provide ablations to verify the effectiveness of various HOTFormerLoc components on Oxford, CS-Campus3D, and \CSWildPlaces. Disabling relay tokens results in a $2.5\%-4.7\%$ drop in performance across all datasets, highlighting the importance of global feature interactions within HOTFormerLoc.

The importance of pyramid attentional pooling is also clear, as we compare the performance of two pooling methods: GeM pooling~\cite{radenovicFineTuningCNNImage2019} using features from a single pyramid level, and a Pyramid GeM pooling, where GeM descriptors are computed for each pyramid level and aggregated with a linear layer. A $7.3\% - 22.5\%$ drop is seen using GeM pooling, and a $2.8\% - 12.3\%$ drop with Pyramid GeM Pooling.

\noindent
\textbf{Octree Depth and Window Size:}
\cref{tab:octree_depth} shows the effect of different octree depths (analogous to input resolution) and attention window sizes on CS-Campus3D and \CSWildPlaces. Overall, an octree depth of 7 with attention window size 64 produces best results. Interestingly, increasing octree depth beyond 7 does not improve performance, which we attribute to redundancy in deeper octrees. A depth of 7 is the minimum needed to represent the $0.8~m$ voxel resolution of \CSWildPlaces~submaps, thus higher depths add no further detail. We see larger attention windows generally improve performance, but not for an octree of depth 6, likely due to sparsely distributed windows in coarser octrees.

\begin{table}
  \centering
  \resizebox{0.8\linewidth}{!}{
  \begin{tabular}{cc cc cc}
    \toprule
     & & CS-Campus3D & \CSWildPlaces \\
    Octree Depth & Window Size & AR@1 $\uparrow$ & AR@1 (Mean) $\uparrow$ \\
    \midrule
    
    \multirow{2}{0 em}{\large 6} & 48 & 79.1 & 36.6 \\
    & 64 & 76.1 & 34.5 \\
    \midrule
    \multirow{2}{0 em}{\large 7} & 48 & 78.4 & 57.6 \\
    & 64 & \textbf{80.4} & \textbf{60.5} \\
    \midrule
    \multirow{2}{0 em}{\large 8} & 48 & 77.0 & 55.9 \\
    & 64 & 79.9 & 58.3 \\
    \bottomrule
  \end{tabular}
  }
  \vspace{-2mm}
  \caption{\small{Ablation study of octree depth and attention window size on CS-Campus3D and \CSWildPlaces.}}
  \label{tab:octree_depth}
  \vspace{-3mm}
\end{table}

\begin{table}
  \centering
  \resizebox{0.35\textwidth}{!}{
  \begin{tabular}{lc cc c cc}
    \toprule
     && \multicolumn{2}{c}{Karawatha} && \multicolumn{2}{c}{Venman} \\
     && AR@1 $\uparrow$ & MRR $\uparrow$ && AR@1 $\uparrow$ & MRR $\uparrow$ \\
    \midrule
    Cartesian && 55.0 & 69.6 && 66.3 & 78.0  \\
    Cylindrical && \textbf{69.6} & \textbf{80.1} && \textbf{80.1} & \textbf{87.4} \\
    \bottomrule
  \end{tabular}
  }
  \vspace{-2mm}
  \caption{\small{Ablation study considering Cartesian vs cylindrical octree attention windows on Wild-Places~\cite{knightsWildPlacesLargeScaleDataset2023}. %
  }}
  \label{tab:ablation_cylindrical}
  \vspace{-5mm}
\end{table}

\noindent
\textbf{Cylindrical Octree Attention Windows:}
\cref{tab:ablation_cylindrical} demonstrates that cylindrical octree attention windows are essential for ground-captured lidar scans in natural environments, contributing to a significant improvement in AR@1 and MRR of $14.6\%$ and $10.5\%$ on Karawatha, and $13.8\%$ and $9.4\%$ on Venman, compared to Cartesian octree attention windows. We note that cylindrical
attention windows best represent point clouds captured by a spinning lidar from the ground, which have a circular pattern. Cartesian attention windows are better suited to the point clouds in Oxford RobotCar~\cite{maddernYear1000Km2017}, which are generated by aggregating consecutive 2D pushbroom lidar scans.

\section{Conclusion and Future Work}
\label{sec:conclusion}
\vspace{-1mm}
We propose \HOTFormerLoc, a novel 3D place recognition method that leverages octree-based transformers to capture multi-granular features through both local and global interactions. We introduce and discuss a new cross-source LPR benchmark,~\CSWildPlaces, designed to advance research on re-localisation in challenging settings.~Our method demonstrates superior performance on our \CSWildPlaces~dataset and outperforms existing SOTA on LPR benchmarks for both ground and aerial views.~Despite these advancements, cross-source LPR remains a promising area for future research. There remain avenues to improve \HOTFormerLoc, such as token pruning to reduce redundant computations and enhancing feature learning with image data that we regard as future work.

\section*{Acknowledgements}
We acknowledge support of the Terrestrial Ecosystem Research Network (TERN), supported by the National Collaborative Infrastructure Strategy (NCRIS). This work was partially funded through the CSIRO's Digital Water and Landscapes initiative (3D-AGB project). We thank the Research Engineering Facility (REF) team at QUT for their expertise and research infrastructure support and Hexagon for providing SmartNet RTK corrections for precise surveying.

{
    \balance
    \small
    \bibliographystyle{ieeenat_fullname}
    \bibliography{main}
}

\clearpage
\setcounter{page}{1}
\maketitlesupplementary
\nobalance

In this document, we present supplementary results and analyses to complement the main paper. \cref{supp:complexity_analysis} provides a complexity analysis of HOTFormerLoc, \cref{supp:cyl_oct_attn} provides visualisations of our cylindrical octree attention, and \cref{supp:pyramid_attn_pool,supp:param_ablations} provide ablations of our pyramidal pooling and network size. \cref{sec:supp_limitations} addresses the limitations and potential future work of our method. 
We include visualisations of our \CSWildPlaces{} dataset in  \cref{sec:supp_dataset_details}. Qualitative examples highlighting components of \HOTFormerLoc, and analysis of the learned attention patterns supported by visualisations are presented in \cref{supp:attention_maps,supp:window_viz}.

\section{\HOTFormerLoc~Additional Details}

\subsection{Complexity Analysis}
\label{supp:complexity_analysis}
Here, we provide a complexity analysis of the components introduced in \cref{sec:hierarchical_octree_attention} of the paper. The key to the efficiency of our approach is alleviating the $O(N^2C)$ complexity of full attention, which is intractable for point clouds with large values of $N$,~\eg~$30K$. This number of points is essential to capture distinctive information in forest environments.
Our H-OSA layer computes windowed attention between non-overlapping windows of size $k$ and their corresponding relay tokens, reducing the complexity to $ O((k+1)^2 \frac{N}{k} C) $. To facilitate global attention at reduced cost, we conduct RTSA on the relay tokens from $L$ levels of the feature pyramid, with complexity $ O(L\frac{N^2}{k^2} C) $. 

Our HOTFormer block thus has a total cost of $ O(L (k+1)^2\frac{N}{k}C + L\frac{N^2}{k^2} C) $. This reduces the quadratic cost relative to $N$ by a factor of $k^2$, but this effect diminishes when $N \gg k$. For this reason, we opt to employ HOTFormer blocks after first processing and downsampling the  $N$ input points into $N_d$ octants with the convolution embedding stem and a series of OSA transformer blocks (similar to H-OSA but with relay tokens disabled), where $N_d < N$. This approach allows us to efficiently initialise strong local features in early stages when semantic information is less developed, which can then be refined by HOTFormer blocks once the size of $N$ is less prohibitive. Another approach would be to consider a larger $k$ for HOTFormer blocks at the finest resolution where $N$ is largest, and smaller values of $k$ at coarser levels, but in this study we have elected to keep $k$ constant throughout the network.

\subsection{Cylindrical Octree Attention}
\label{supp:cyl_oct_attn}
\begin{figure}[t]
  \centering
   \setlength{\fboxsep}{0pt}%
   \setlength{\fboxrule}{.5pt}%
   \fbox{
   \includegraphics[width=.99\linewidth]{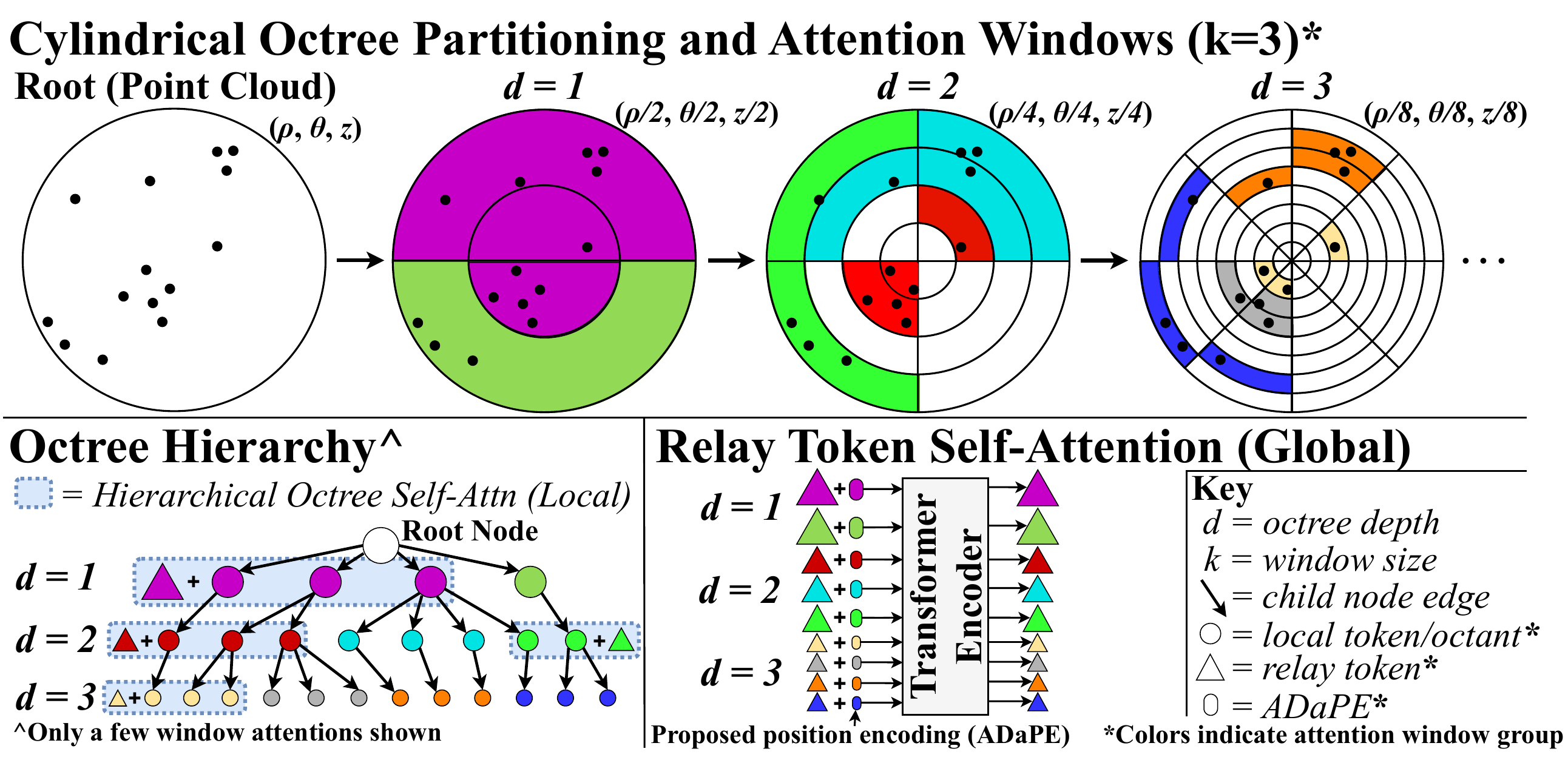}
   }
   \vspace{-5mm}
   \caption{\small{Cylindrical Octree Hierarchy and proposed attention mechanisms shown in 2D for simplicity (3D extends with $z$-axis, so technically the above is a quadtree). Cylindrical partitions and tree nodes are color-matched.}}
   \label{fig:supp_cylindrical_attention_hierarchy}
   \vspace{-5mm}
\end{figure}

In \cref{fig:supp_cylindrical_attention_hierarchy} we visualise the relationship between the cylindrical octree hierarchy (albeit in 2D) up to depth 3, and corresponding attention windows with window size $k=3$ (grouped by color) following $z$-ordering as described in \cref{sec:cylindrical_octrees}. The HOTFormerLoc structure detailed in ~\cref{fig:hotformer_overview} can be used interchangeably with Cartesian or cylindrical octree attention windows.

\begin{table}[b]
  \centering
  \vspace{-3mm}
  \resizebox{1.0\linewidth}{!}{
  \begin{tabular}{c c c c}
    \toprule
     & Oxford & CS-Campus3D & \CSWildPlaces \\
    Pooled Tokens & AR@1 (Mean) $\uparrow$ & AR@1 $\uparrow$ & AR@1 (Mean) $\uparrow$ \\
    \midrule
    74, 36, 18 & \textbf{92.1} & 79.8 & \textbf{60.5} \\
    148, 72, 36 & 91.1 & \textbf{80.4} & 52.7 \\
    296, 144, 72 & 89.8 & 74.9 & 48.4 \\
    \bottomrule
  \end{tabular}
  }
  \vspace{-2mm}
  \caption{\small{Ablation study considering the number of pooled tokens used for pyramid attentional pooling on Oxford, CS-Campus3D and \CSWildPlaces.}}
  \label{tab:supp_attn_pool}
  \vspace{-6mm}
\end{table}

\subsection{Pyramid Attentional Pooling}
\label{supp:pyramid_attn_pool}
We provide an ablation of our pyramid attention pooling (proposed in \cref{sec:pyramid_attn_pool}) in \cref{tab:supp_attn_pool}, using different numbers of pooled tokens $q$ on Oxford~\cite{maddernYear1000Km2017}, CS-Campus3D~\cite{guanCrossLoc3DAerialGroundCrossSource2023} and \CSWildPlaces. Overall, we find $q=[74,36,18]$ to produce the best results across most datasets, although $q=[148,72,36]$ performs marginally better on CS-Campus3D. 

These multi-scale pooled tokens $\Omega_l$ are concatenated to form $\Omega'$ and processed by the token fuser~\cite{ali-beyMixVPRFeatureMixing2023}, generating $q_{\mathrm{total}}=128$ tokens with $C=256$ channels in our default configuration. In the MLP-Mixer~\cite{tolstikhinMLPMixerAllMLPArchitecture2021}, the channel-mixing and token-mixing MLPs project these tokens to $\bar{k}=32$ and $\bar{C}=8$, which are then flattened and $L_2$-normalised to produce the 256-dimensional global descriptor $d_\mathcal{G}$.

\begin{table}[t]
  \centering
  \resizebox{1\linewidth}{!}{
  \begin{tabular}{@{}ccccccc@{}}
    \hline
    &&& Runtime & Oxford & CS-Campus3D & \CSWildPlaces{} \\
    Channels & Blocks & Params & (Sparse / Dense) & AR@1 (Mean) & AR@1 & AR@1(Mean)\\
    \hline
    C = 256 & M = 10 & 35.4 M & 62 / 270 ms & 92.1 \textcolor{blue}{($\uparrow$2.1)} & 80.4 \textcolor{blue}{($\uparrow$9.7)} & 60.5 \textcolor{blue}{($\uparrow$8.5)} \\
    C = 256 & M = 8 & 28.9 M & 50 / 250 ms & 91.8 \textcolor{blue}{($\uparrow$1.8)}  & 75.5 \textcolor{blue}{($\uparrow$4.8)} & 58.9 \textcolor{blue}{($\uparrow$6.9)} \\
    C = 256 & M = 6 & 22.6 M & 41 / 228 ms & 91.5 \textcolor{blue}{($\uparrow$1.5)} \ & 71.9 \textcolor{blue}{($\uparrow$1.2)} & 57.6 \textcolor{blue}{($\uparrow$5.6)} \\
    C = 192 & M = 8 & 16.7 M & 40 / 192 ms & 90.8 \textcolor{blue}{($\uparrow$0.8)} & 75.2 \textcolor{blue}{($\uparrow$4.5)} & 58.1 \textcolor{blue}{($\uparrow$6.1)} \\
    \hline
  \end{tabular}
  }
  \vspace{-2mm}
  \caption{\small{Ablation on number of HOTFormer blocks and channel size. \textcolor{blue}{($\uparrow$X.X)} indicates improvement in AR@1 over SOTA method per-dataset.}}
  \label{tab:supp_num_blocks_ablation}
  \vspace{-5mm}
\end{table}

\subsection{HOTFormerLoc Ablations}
\label{supp:param_ablations}
We provide ablations on the number of HOTFormer blocks and channel size in \cref{tab:supp_num_blocks_ablation}. HOTFormerLoc maintains SOTA performance with fewer parameters than the full-sized model, outperforming MinkLoc3Dv2 by $22.7\%$ on CS-Campus3D and $6.1\%$ on CS-Wild-Places with just 16.7M params. This parameter count is similar to existing transformer-based LPR methods~\cite{xuTransLoc3DPointCloud2022,huiPyramidPointCloud2021}, whilst outperforming them by $32.2\%$ on CS-Campus3D. We also report the runtime on dense point clouds from \CSWildPlaces{}, and the sparse point clouds from CS-Campus3D, with HOTFormerLoc achieving $40-62~ms$ inference time when limited to 4096 points.

\subsection{Limitations and Future Work}
\label{sec:supp_limitations}
While \HOTFormerLoc~has demonstrated impressive performance across a diverse suite of LPR benchmarks, it has some limitations. The processing of multi-grained feature maps in parallel is a core design of HOTFormerLoc, and while effective, it causes some redundancy. For example, there is likely a high correlation between features representing the same region in different levels of the feature pyramid. Currently, these redundant features can be filtered by the pyramid attentional pooling layer, but this does not address the wasted computation earlier in the network within HOTFormer blocks. In future work, token pruning approaches can be adopted to adaptively remove redundant tokens, particularly at the finest resolution where RTSA is most expensive to compute.

Another source of redundancy is related to the number of parameters in our network. A large portion of these are attributed to the many transformer blocks, as each pyramid level has its own set of H-OSA layers with channel size $C=256$. In the future, the parameter count can be reduced by utilising different channel sizes in each level of the feature pyramid, with linear projections to align the dimensions of relay tokens during RTSA. 

As mentioned in \cref{sec:runtime_analysis}, the runtime of HOTFormerLoc can be improved through parallelisation. While our design is best suited for parallel implementation, currently, the H-OSA layers for each pyramid level are computed in serial. To unlock the full potential of our network design for optimal runtime, these layers can be combined into a single operation. Furthermore, the octree implementation used in HOTFormerLoc can be parallelised to enable more efficient octree construction.

\section{\CSWildPlaces~Dataset Visualisations}
\label{sec:supp_dataset_details}
\begin{figure}[t]
  \centering
   \includegraphics[width=\linewidth]{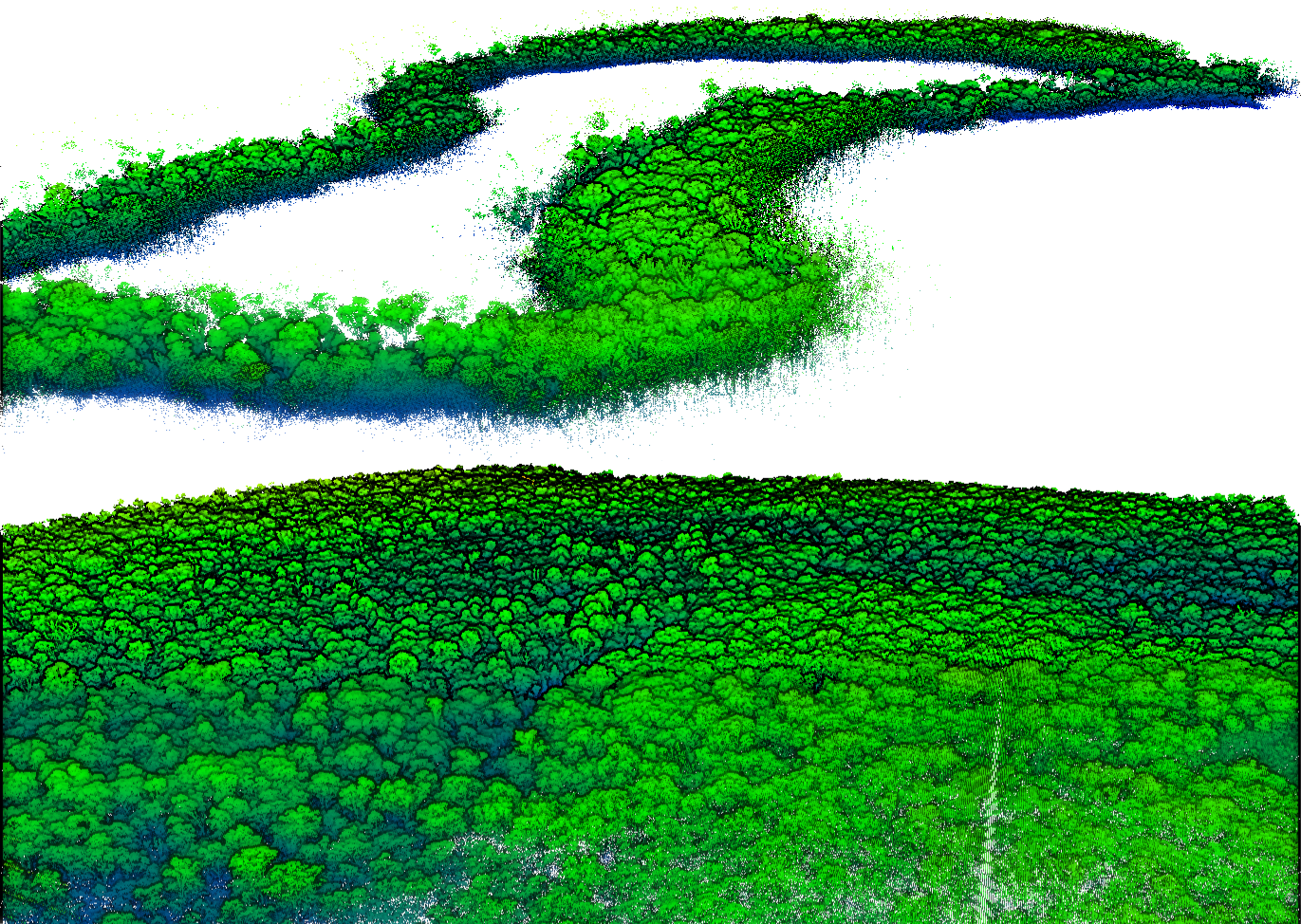}
    \vspace{-4mm}
    \caption{\small{Matched portions of the ground (top) and aerial (bottom) global maps from Karawatha forest in \CSWildPlaces. The aerial maps cover a significantly larger area than the ground traversals, increasing the likelihood of false positive retrievals. Maps are shifted along z for visualisation purposes.
    }}
    \label{fig:supp_cswildplaces_stacked}
    \vspace{-5mm}
\end{figure}

We provide additional visualisations of our \CSWildPlaces~dataset, highlighting its unique  characteristics. In \cref{fig:supp_cswildplaces_stacked}, we compare a section of the ground and aerial global maps from Karawatha. One notable feature of our dataset is the large-scale aerial coverage, creating a challenging retrieval task where ground queries must be matched against potentially tens of thousands of candidates.

\begin{figure*}[t]
  \centering
  \begin{subfigure}{0.494\linewidth}
   \includegraphics[width=\linewidth,clip]{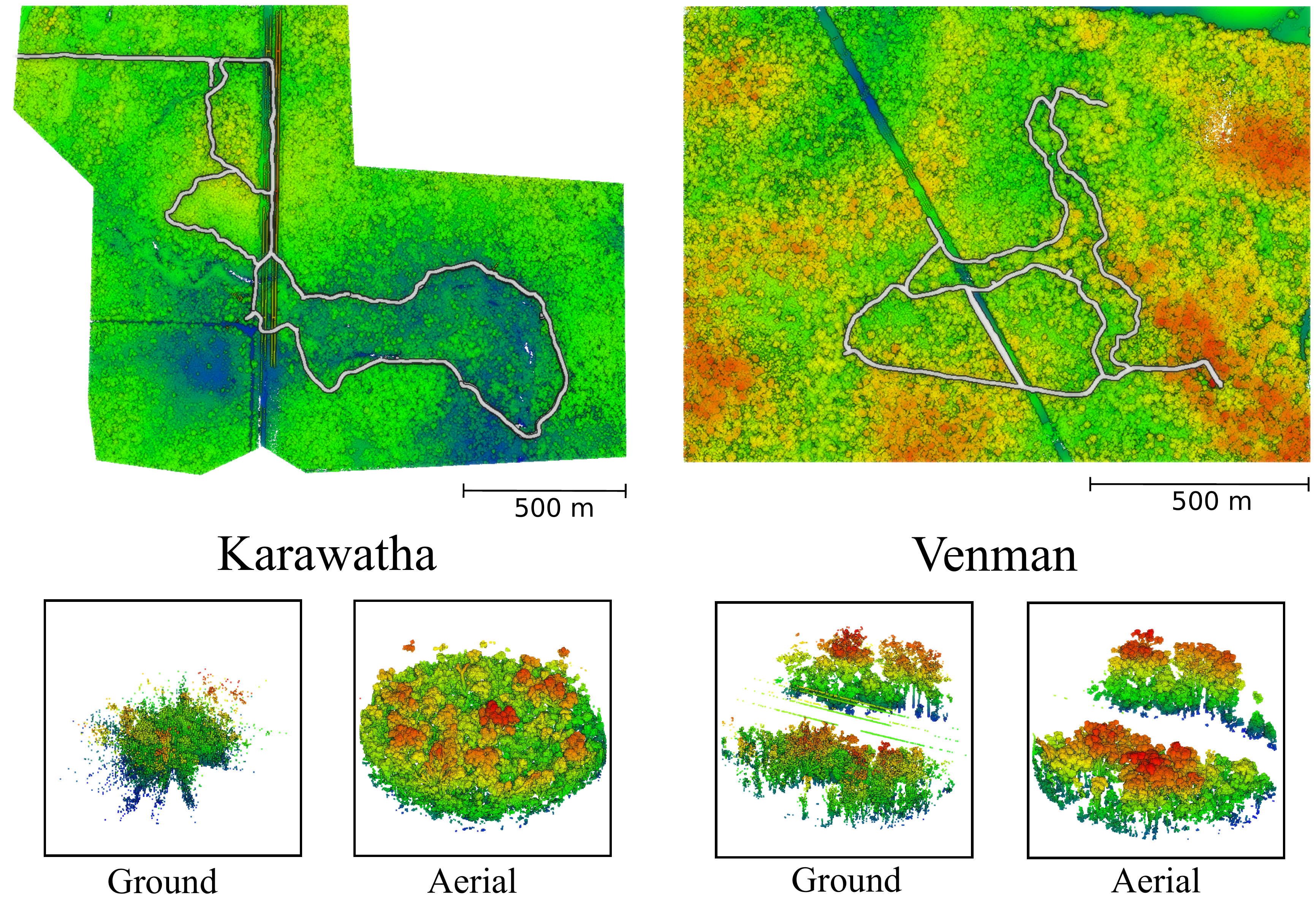}
   \caption{\small{Baseline forests}}
  \end{subfigure}
  \hfill
  \begin{subfigure}{0.496\linewidth}
   \includegraphics[width=\linewidth]{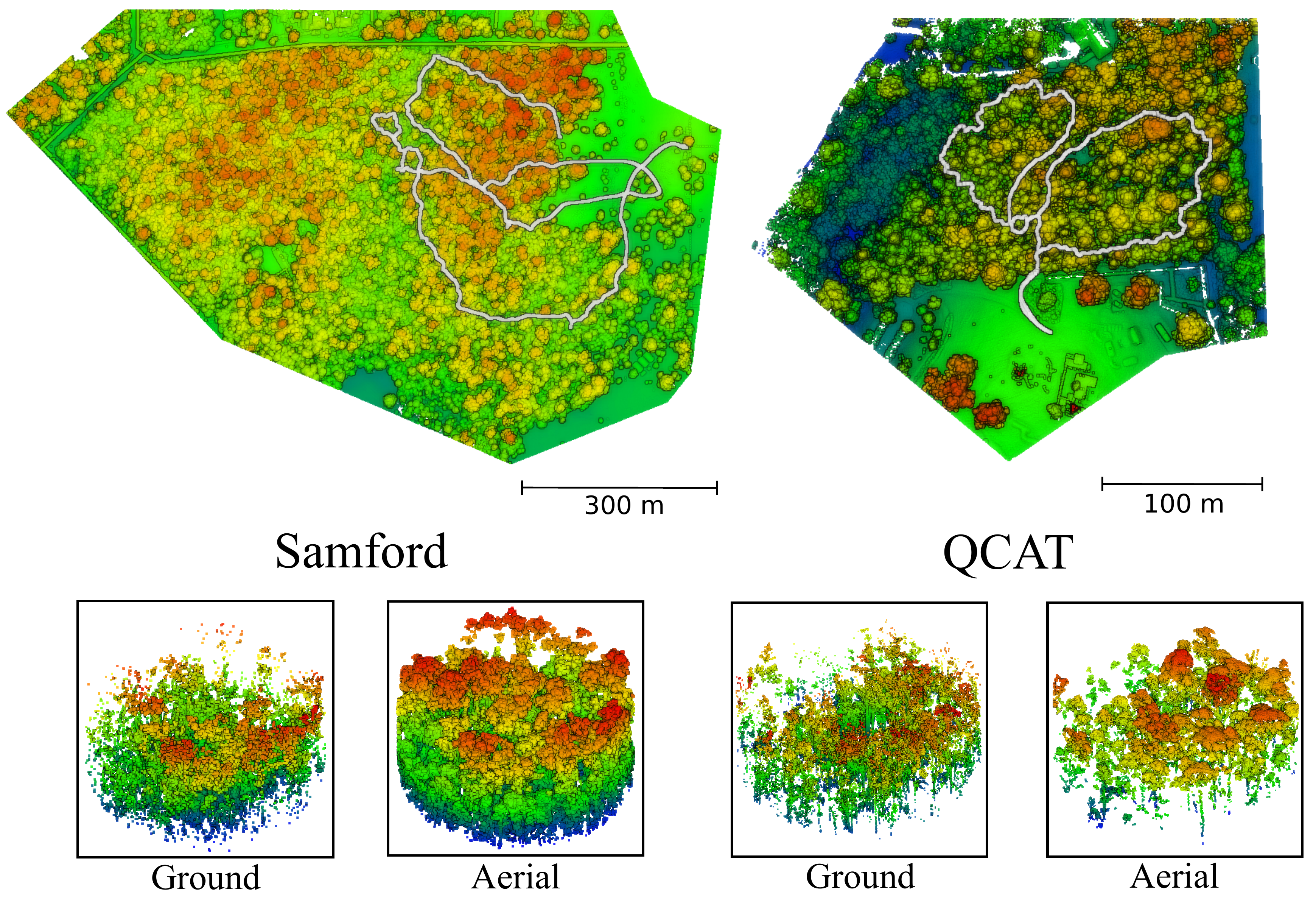}
   \caption{\small{Unseen forests}}
  \end{subfigure}
    \vspace{-2mm}
    \caption{\small{(Top row) bird's eye view of aerial maps from all forests of \CSWildPlaces. (Bottom row) ground and aerial submap from each. Our dataset features high-resolution ground and aerial lidar scans from four diverse forests, with major occlusions between viewpoints.
    }}
    \label{fig:supp_cswildplaces_overview}
    \vspace{-5mm}
\end{figure*}

In \cref{fig:supp_cswildplaces_overview}, we exhibit the scale and point distribution of all four forest environments in the \CSWildPlaces~dataset. The Baseline forests have a combined aerial coverage of $3.1~km^2$, while the Unseen forests add a further $0.6~km^2$ of aerial coverage. Submaps visualised from each forest showcase the distinct distributional differences between environments. Additionally,  the limited overlap between ground and aerial perspectives clearly demonstrate why ground-to-aerial LPR in forested areas is challenging. Notably, our dataset is the first to provide high-resolution aligned aerial and ground lidar scans of this scale in forested environments, offering a valuable benchmark for training and evaluating place recognition approaches. 

\section{Attention Map Visualisations}
\label{supp:attention_maps}
\begin{figure}[t]
  \centering
  \begin{subfigure}{0.99\linewidth}
      \includegraphics[width=.99\linewidth,trim={0 0 0 1.0cm},clip]{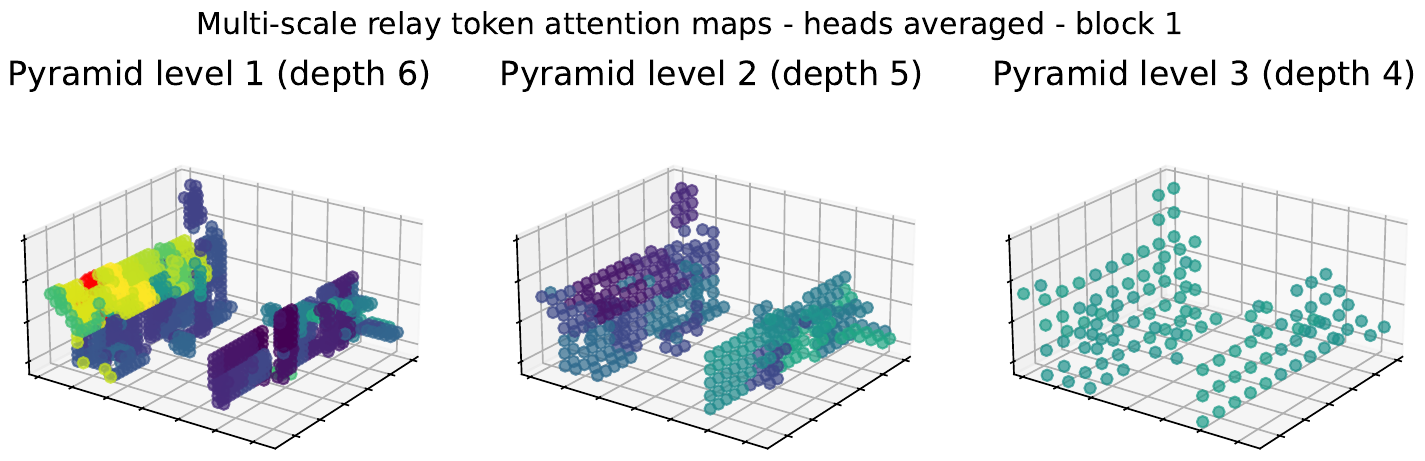}
      \caption{First RTSA block}
  \end{subfigure}
  \hfill  
  \begin{subfigure}{0.99\linewidth}
      \includegraphics[width=.99\linewidth,trim={0 0 0 2.5cm},clip]{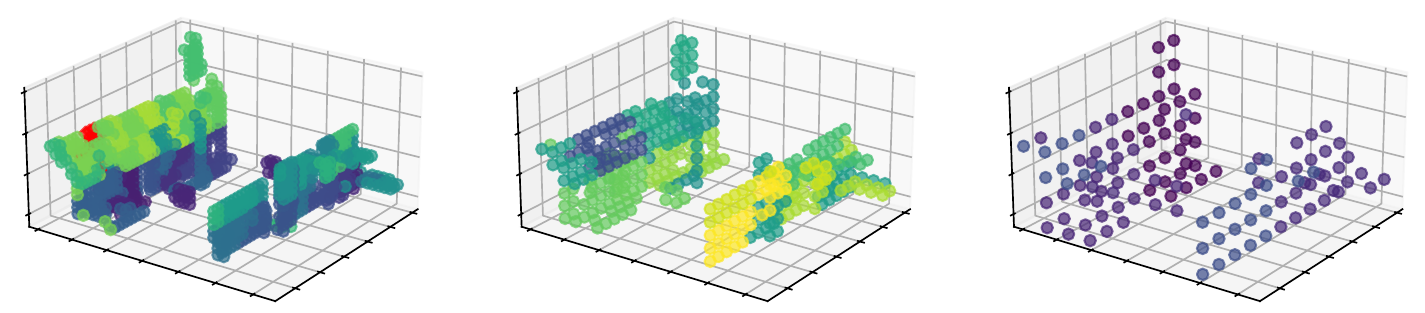}
      \caption{Mid RTSA block}
  \end{subfigure}
  \hfill  
  \begin{subfigure}{0.99\linewidth}
      \includegraphics[width=.99\linewidth,trim={0 0 0 2.5cm},clip]{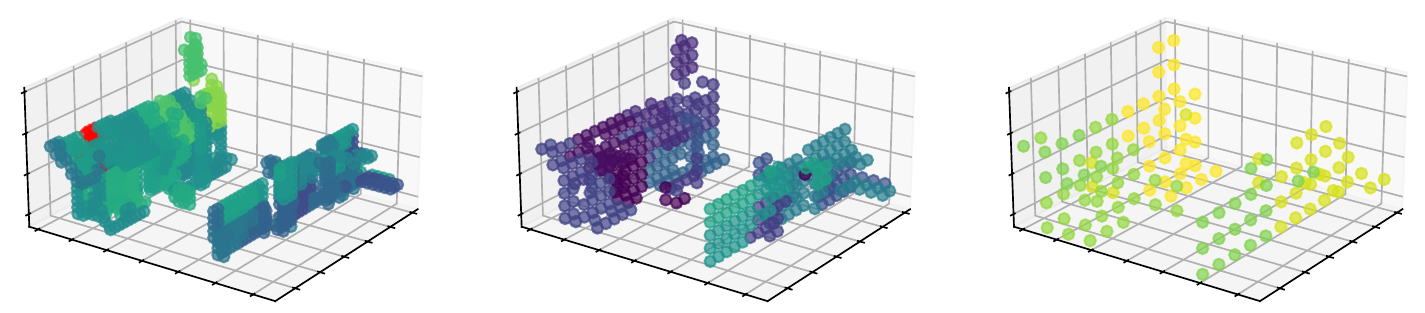}
      \caption{Last RTSA block}
  \end{subfigure}
  \vspace{-2mm}
  \caption{\small{Relay token multi-scale attention visualised on the octree feature pyramid at different layers in the network, colourised by attention weight relative to the red query token (brighter colours indicate higher weighting). The network learns a local-to-global attention pattern from the first to last layer.}}
  \label{fig:supp_hierarchical_attn}
  \vspace{-5mm}
\end{figure}

\begin{figure*}[t]
    \begin{minipage}{0.49\linewidth}
        \centering
        \includegraphics[width=\textwidth,trim={0.8cm 0.5cm 0.8cm 1.4cm},clip]{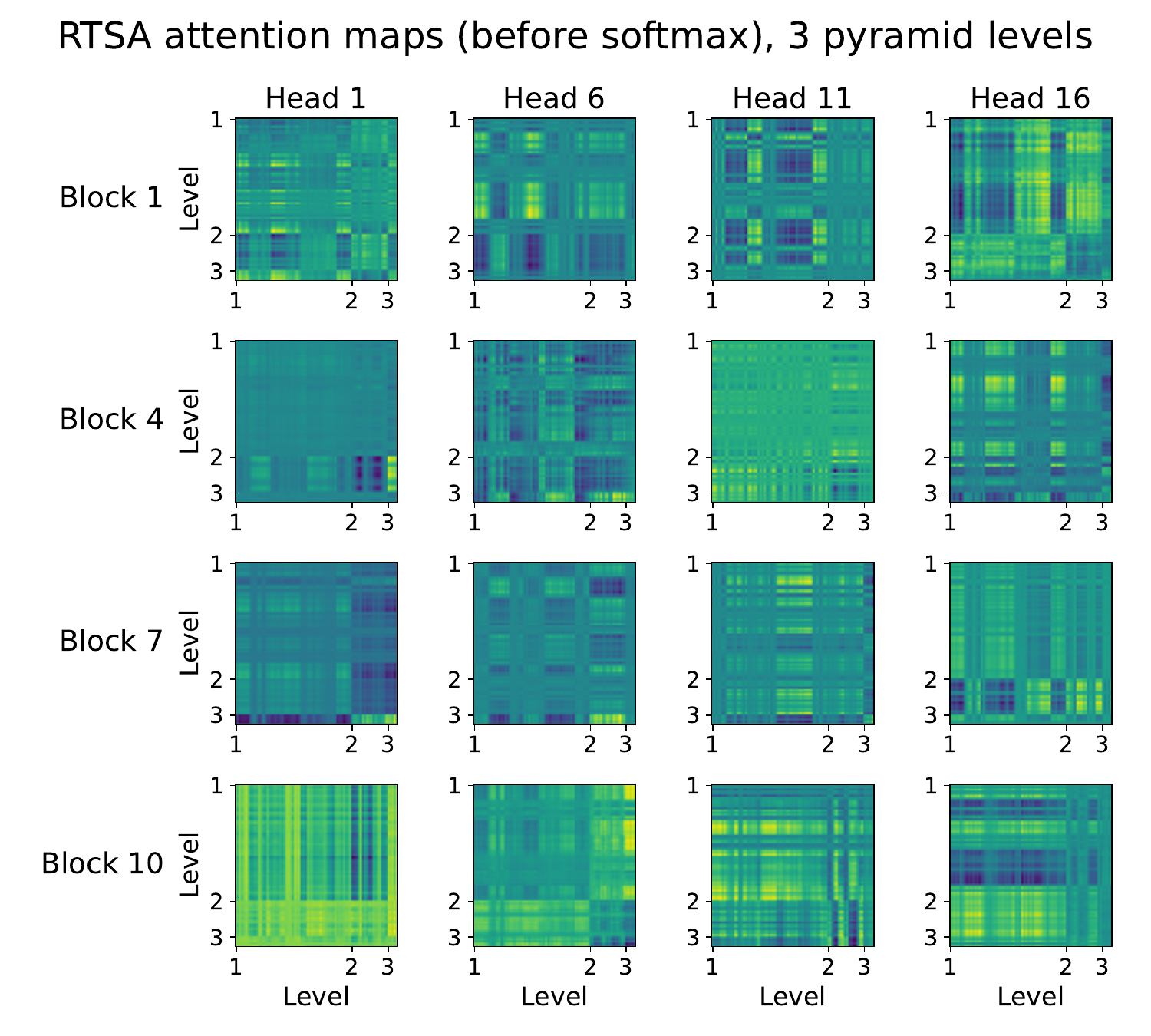}
        \caption{Multi-scale relay token attention matrices from different RTSA heads and blocks for a submap from Oxford. Attention heads learn to focus on different feature granularities (axis ticks indicate pyramid level of corresponding relay tokens). }
        \label{fig:supp_rt_attn_map}  
    \end{minipage}
    \hfill
    \begin{minipage}{0.49\linewidth}
        \centering
        \includegraphics[width=0.99\linewidth,trim={0.6cm 0.5cm 0.6cm 1.4cm},clip]{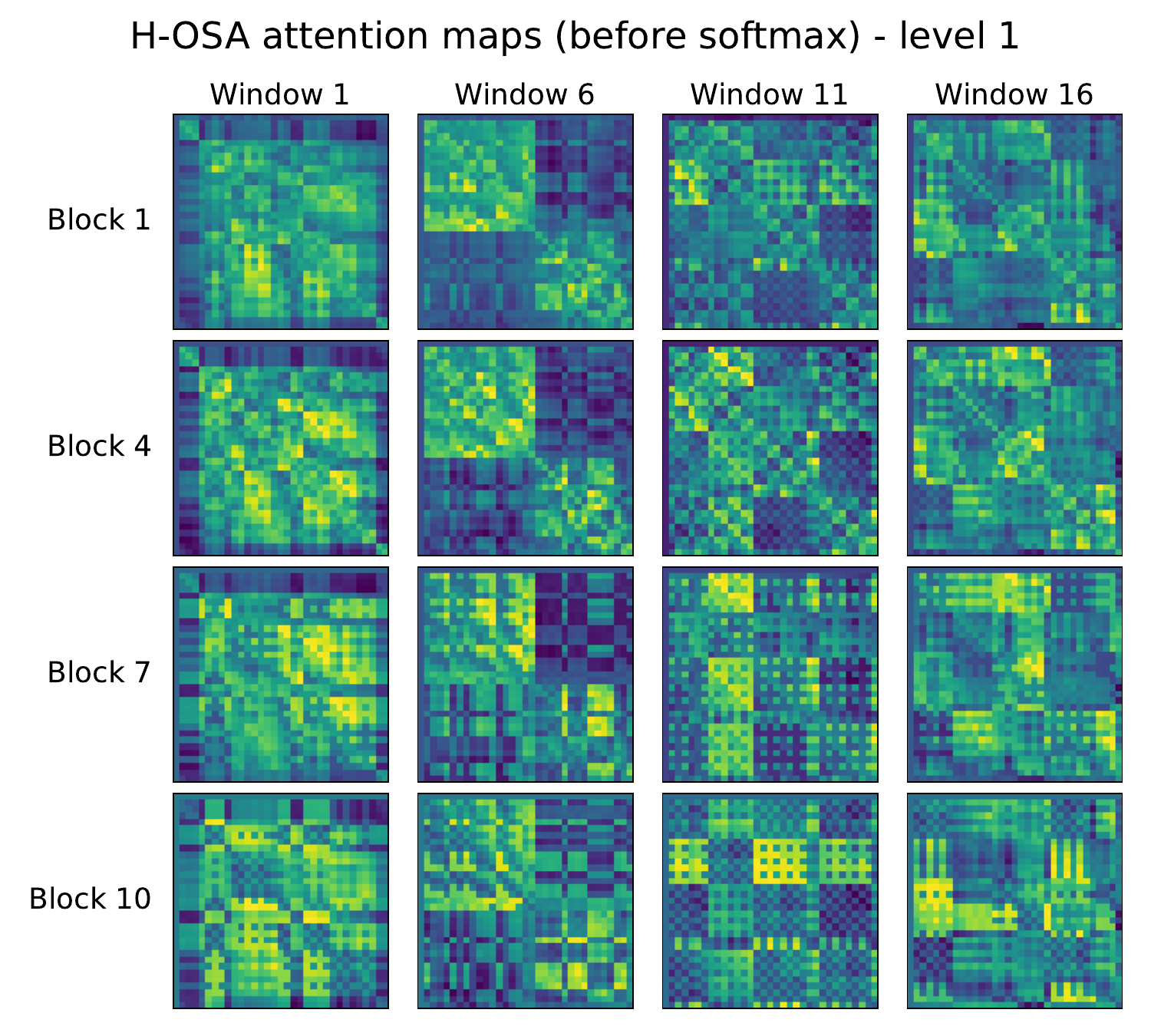}
        \caption{\small{Local attention matrices from different attention windows within H-OSA blocks (averaged over attention heads) for a submap from Oxford. The relay token is represented by the top-left element of each map.}}
        \label{fig:supp_hosa_attn_map}  
    \end{minipage}
\end{figure*}

We provide visualisations of the local and global attention patterns learned by HOTFormerLoc in \cref{fig:supp_hierarchical_attn,fig:supp_hosa_attn_map,fig:supp_rt_attn_map}. In \cref{fig:supp_hierarchical_attn}, we analyse the attention patterns learnt by RTSA for a submap from the Oxford dataset~\cite{maddernYear1000Km2017} to verify the intuition behind relay tokens. Here, we visualise the attention scores of the multi-scale relay tokens within the octree representation for each level of the feature pyramid (where points represent the centroid of each octant, for ease of visualisation). We select a query token (highlighted in red), and colourise other tokens in all pyramid levels by how strongly the query attends to each (yellow for strong activation, purple for weak activation). We compare the attention patterns of this query token from the first, middle, and last RTSA layer in the network. 

We see that RTSA learns a local-to-global attention pattern as it progresses through the network. In the first RTSA layer, the query token primarily attends to other neighbour tokens of the same granularity. In the middle RTSA layer, the local neighbourhood is still highly attended to, but we see higher attention to distant regions in level 2 of the feature pyramid with coarser granularity. In the final layer, the query token primarily attends to tokens in the coarsest level of the pyramid, taking greater advantage of global context. We provide further visualisations of the attention matrices from RTSA in \cref{fig:supp_rt_attn_map}, which highlights the multi-granular attention patterns learnt by different attention heads as tokens propagate through the HOTFormer blocks.

In \cref{fig:supp_hosa_attn_map}, we visualise the attention patterns of H-OSA layers, comparing the patterns learnt for different local attention windows as tokens pass through each HOTFormer block. In particular, the presence of strong local dependencies is indicated by square regions with high activations. Interestingly, the relay token (top- and left-most element of each matrix) is uniformly attended to by the local tokens in each window, but with gradually higher attention values in later HOTFormer blocks, indicating the shift towards learning global context in later stages of the network. 

\section{Octree Attention Window Visualisations}
\label{supp:window_viz}
\begin{figure*}[b]
    \centering
    \vspace{-2mm}
    \begin{subfigure}[b]{0.24\linewidth}
        \centering
        \includegraphics[width=\textwidth]{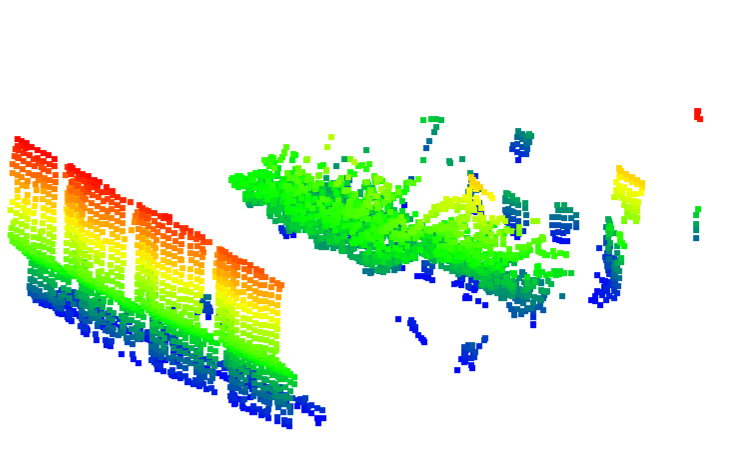}
        \caption{\small{Submap}}
    \end{subfigure}
    \hfill
    \begin{subfigure}{0.75\linewidth}
        \includegraphics[width=\linewidth,trim={3cm 2.8cm 2.2cm 1.5cm},clip]{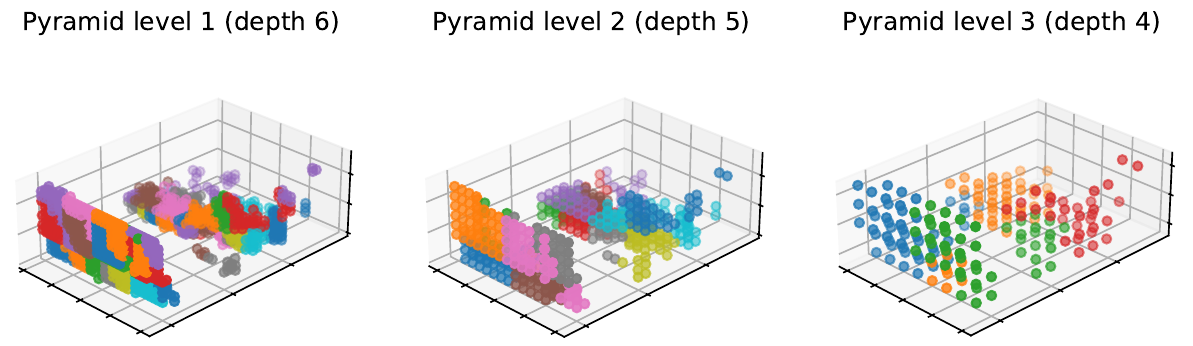}
        \caption{\small{Cartesian attention windows}}
    \end{subfigure}
    \vfill
    \begin{subfigure}[b]{0.24\linewidth}
    \end{subfigure}
    \hfill
    \begin{subfigure}{0.75\linewidth}
        \includegraphics[width=\linewidth,trim={3cm 2.8cm 2.2cm 3cm},clip]{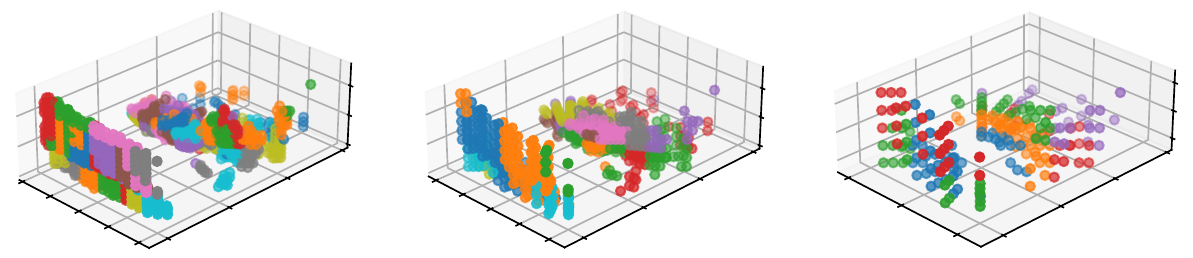}
        \caption{\small{Cylindrical attention windows}}
    \end{subfigure}
    \caption{\small{Comparison of Cartesian \vs cylindrical octree attention windows on submaps from Oxford Robotcar~\cite{maddernYear1000Km2017}, where nearby points are colourised by which local attention window they belong to. The uniform nature of aggregated 2D lidar scans and highly-structured scene geometry make Cartesian attention windows a better representation for Oxford.
    }}
    \label{fig:supp_cart_windows}
    \vspace{-3mm}
\end{figure*}

\begin{figure*}[pt]
    \centering
    \vspace{-2mm}
    \begin{subfigure}[b]{0.22\linewidth}
        \centering
        \includegraphics[width=\textwidth,trim={6cm 0cm 6cm 1.5cm},clip]{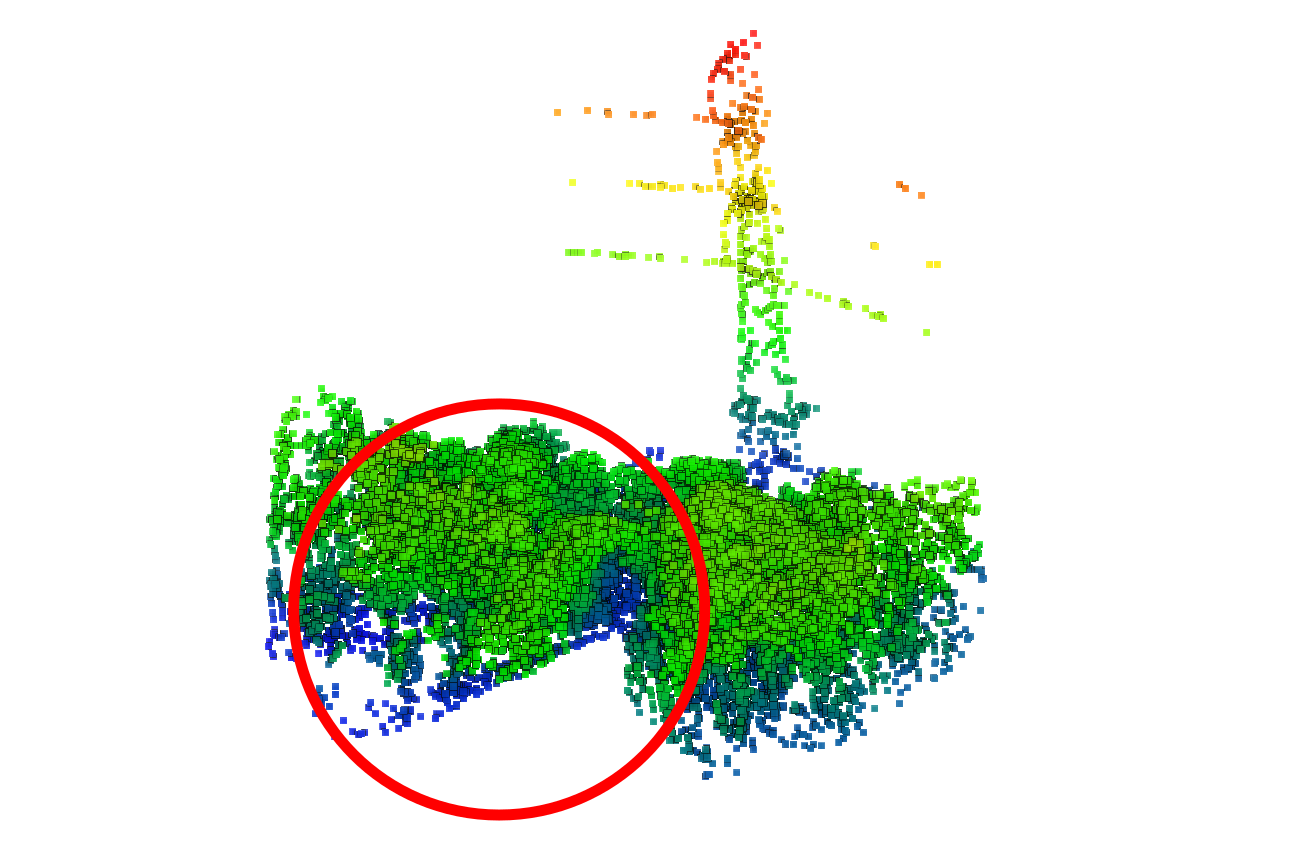}
        \caption{\small{Submap}}
    \end{subfigure}
    \hfill
    \begin{subfigure}{0.77\linewidth}
        \centering
        \includegraphics[width=\textwidth,trim={3.2cm 2.3cm 2.5cm 1.5cm},clip]{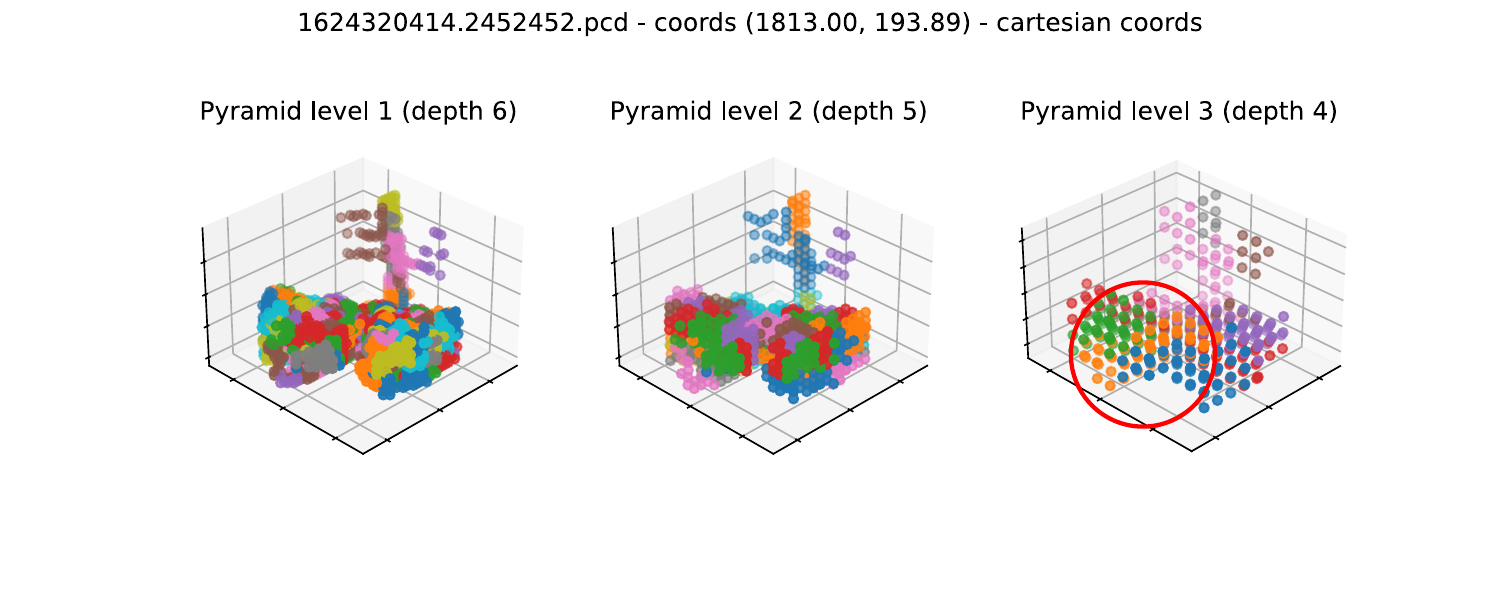}
        \caption{\small{Cartesian attention windows}}
    \end{subfigure}
    \vfill
    \begin{subfigure}[b]{0.24\linewidth}
    \end{subfigure}
    \hfill
    \begin{subfigure}{0.77\linewidth}
        \centering
        \includegraphics[width=\textwidth,trim={3.2cm 2.3cm 2.5cm 2.5cm},clip]{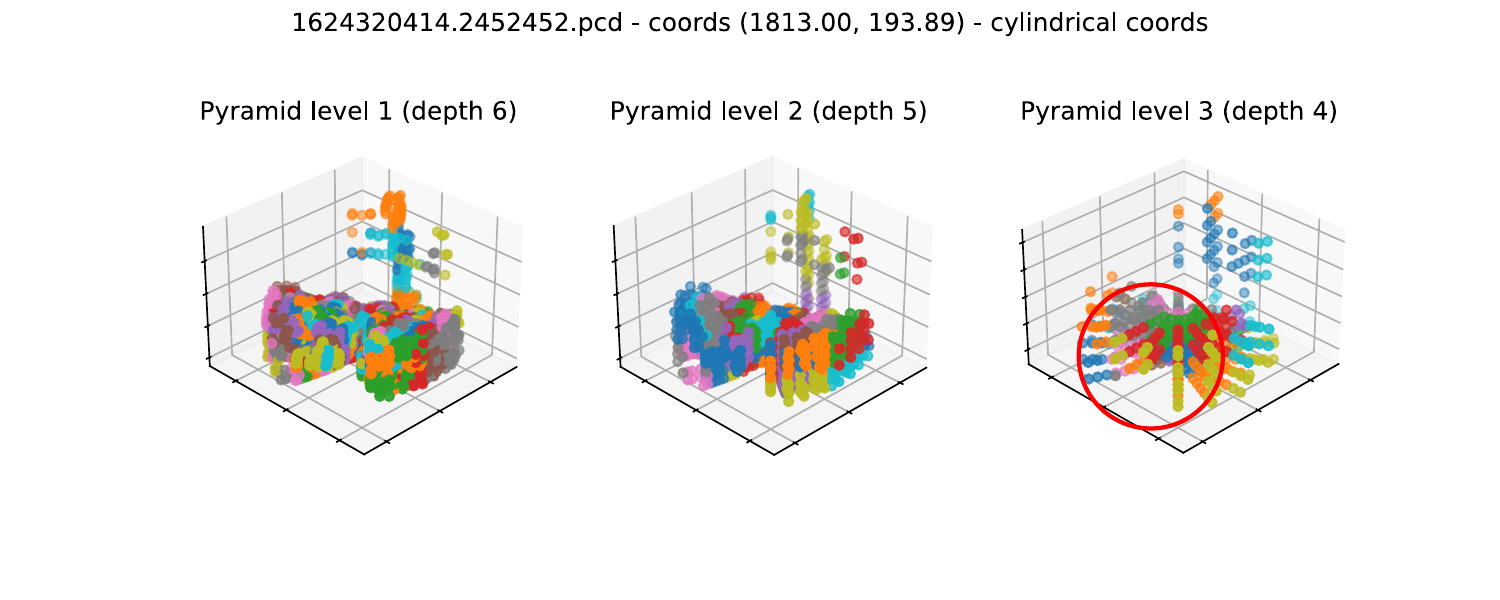}
        \caption{\small{Cylindrical attention windows}}
    \end{subfigure}    
    \caption{\small{Comparison of Cartesian \vs cylindrical octree attention windows on submaps from Wild-Places~\cite{knightsWildPlacesLargeScaleDataset2023}. The variable density of spinning lidar is better captured by cylindrical attention windows in coarser levels, and tree trunks are better represented. We highlight a region where the effect is most noticeable.
    }}
    \label{fig:supp_cyl_windows}
    \vspace*{4in}
\end{figure*}

In \cref{fig:supp_cart_windows} and \cref{fig:supp_cyl_windows} we visualise Cartesian and cylindrical octree attention windows generated on real submaps from Oxford~\cite{maddernYear1000Km2017} and Wild-Places~\cite{knightsWildPlacesLargeScaleDataset2023}. On the Oxford dataset, which features highly structured urban scenes with flat geometries (such as walls), Cartesian octree windows are a better representation of the underlying scene. Point clouds in Oxford are generated by aggregating 2D lidar scans, as opposed to a single scan from a spinning lidar, producing a uniform point distribution. Furthermore, at coarser levels, the cylindrical octree distorts the flat wall on the left side of the scene to appear as though it is curved. For these reasons, we find that Cartesian octree attention windows perform best on this data.

In contrast, we see the advantage of cylindrical octree attention windows on a submap from Wild-Places in \cref{fig:supp_cyl_windows}. In the red circled region, it is clear that the coarsest level of the cylindrical octree better represents the shape and distribution of circular lidar scans than the Cartesian octree. Further, the size of each cylindrical attention window reflects the density of points, with smaller, concentrated windows near the centre, and larger, sparse windows towards the edges of the scene. In contrast, the Cartesian attention windows all cover a similar sized region.

\end{document}